\documentclass{article}

\usepackage[left=1.5in,top=1.5in,right=1.5in,bottom=1.5in,nohead,paperwidth=8.5in, paperheight=11in]{geometry} 

\usepackage{graphicx}
\usepackage[utf8]{inputenc}

\usepackage{natbib}
\setcitestyle{authoryear,open={(},close={)}}

\usepackage{graphicx}
\usepackage{subcaption}
\usepackage{multirow}
\usepackage{amsmath} 
\usepackage{amsfonts} 
\usepackage{stmaryrd}
\usepackage{float}
\usepackage{rotating}
\usepackage{multirow}
\usepackage[titletoc, title]{appendix}
\usepackage{adjustbox}
\usepackage{bbm}
\usepackage{color}
\usepackage[table]{xcolor}
\usepackage{lscape}
\usepackage{tablefootnote}
\usepackage{caption}
\usepackage{morefloats}
\usepackage[utf8]{inputenc}
\usepackage{tikz}
\def\checkmark{\tikz\fill[scale=0.4](0,.35) -- (.25,0) -- (1,.7) -- (.25,.15) -- cycle;} 

\usepackage{array} 
\usepackage{tabularx} 

\usepackage{comment}

\usepackage{stfloats}

\usepackage{url}
\usepackage{wasysym}
\usepackage{soul} 

\usepackage{todonotes} 

\makeatletter
\newif\if@restonecol
\makeatother

\usepackage[ruled,slide]{algorithm2e} 

\usepackage{authblk}

\usepackage[misc]{ifsym}

\title{Neural Forecasting of the Italian Sovereign Bond Market with Economic News\footnote{The views expressed are purely those of the writer and may not in any circumstance be regarded as stating an official position of the European Commission.}} 

\author[1,*]{Sergio Consoli}
\author[1]{Luca Tiozzo Pezzoli}
\author[2]{Elisa Tosetti}
\affil[1]{\small{European Commission, Joint Research Centre (JRC), 
Via E. Fermi 2749, I-21027 Ispra (VA), Italy. \Letter \textit{\{name.surname\}@ec.europa.eu}}}

\affil[2]{\small{Università Ca' Foscari Venezia, Department of Management, Cannaregio 873, Fondamenta San Giobbe, 30121 Venezia, Italy. \Letter \textit{elisa.tosetti@unive.it}}}

\affil[*]{\small{Corresponding author \phone (+39)0332-786584, \Letter \textit{sergio.consoli@ec.europa.eu}}}

\date{}

\begin{document}

\maketitle

\begin{abstract}

In this paper we employ economic news within a neural network framework to forecast the Italian 10-year interest rate spread. We use a big, open-source, database known as Global Database of Events, Language and Tone to extract topical and emotional news content linked to bond markets dynamics. 
We deploy such information within a probabilistic forecasting framework with autoregressive recurrent networks (DeepAR). Our findings suggest that a deep learning network based on Long-Short Term Memory cells outperforms classical machine learning techniques and provides a forecasting performance that is over and above that obtained by using conventional determinants of interest rates alone.

\bigskip
\textbf{Keywords: }{Big data; Government yield spread; GDELT; Machine learning; Neural forecasting; Feature engineering.}

\end{abstract}

\newpage

\section{Introduction} \label{introduction} 
 


Since the global financial crisis in 2008, European government yield spreads have experienced numerous shocks, including the European sovereign debt crisis in 2012 and the recent economic slump experienced with the surge of the COVID-19 pandemic \citep{zhang2020}. In periods of economic distress, traditional determinants of yield spreads 
adopted by existing studies to parsimoniously model and forecast interest rates dynamics may not well summarise investors' information about the creditworthiness of a country \citep{Diebold_Li_06}.
Recent work from the behavioural finance literature highlights the importance of human cognition and sentiment as forces that significantly influence investors' perception, expectations and decisions 
\citep{Blommestein2012}. A number of studies have proposed to exploit the polarity of the language used in social media and in journal articles to approximate the sentiment of investors. In a seminal work, \citet{tetlock2007} extracts investors' sentiment from financial newspapers and uses it to predict financial stock market returns, showing how it significantly enhances the performance of classical predictors, particularly during periods of economic distress \citep{Garcia2013}. More recently, the use of sentiment extracted from text data has been successfully employed to predict bond markets interest rates \citep{BEETSMA2013, Liu2014} and Credit Default Swaps \citep{Apergis2015, Apergis2016}. These studies find that an increase in media pessimism deteriorates the risk profile of a country thus producing upward movements in spreads.


In this paper, we exploit the potentials of the worldwide largest news-based database existing to date, known as the \textit{Global Database of Events, Language and Tone} (GDELT) \citep{Leetaru_2013gDELT} to improve existing forecasting models of the sovereign bond market. GDELT is an open-source platform that collects, translates into English, processes news worldwide, and updates them every fifteen minutes on a dedicated web-platform, publicly available\footnote{GDELT website: \url{https://www.gdeltproject.org/}.}. It extracts a vast number of important features from news, by looking at locations, people and organizations mentioned in the text, and calculating themes and sentiment by means of popular topical taxonomies and dictionaries. 
We propose a simple feature selection procedure to extract from GDELT a set of indicators capturing investors' emotions, sentiments and topics popularity from Italian news and then use them to forecast daily changes in the 10-year Italian interest rate yield against its German counterpart, using data for the period from the 2nd of March 2015 to the 31st of August 2019. %
Spreads measured against Germany are commonly used in the financial literature, where German bonds are considered as the risk-free benchmark asset for Europe \citep{Afonso_Arghyrou_Kontonikas_2015,Arghyrou_Kontonikas_2012}. Therefore, Italian spreads relative to Germany can be seen as the compensation demanded by investors for taking the additional risk relative to an investment in the safer German bonds. 

The typical statistical model adopted to forecast sovereign government bond spreads is a linear regression, possibly incorporating time dependency \citep{Baber_Brandt_Kavajecz_2009,Favero_2013,Liu2014}. While such assumption considerably simplifies the analysis, it may not be reliable when incorporating in the model information extracted from alternative, large databases, where extracted features are often highly correlated and carry low signals. 
In addition, several studies have shown that during periods of financial distress, complex non-linear relationships among explanatory variables may affect the behaviour of the sovereign bond spreads distribution, at specific quantiles, that simple linear models are not able to capture \citep{Bernal_Gnabo_Guilmin_2016}. The sample period examined in this paper is characterised by high political and economic uncertainty mainly due to the difficulty in forming a new Italian government in early summer 2018 and the debate about deficit spending engagements with the European Union later that year.
To account for these characteristics and assess the predictive power of the selected GDELT variables over traditional determinants, in this paper we adopt the probabilistic deep neural network approach by \citet{Salinas2020}, known as DeepAR. 
This approach builds upon previous work on deep learning for time series data and consists of fitting an artificial recurrent neural network model with feedback connections \citep{Schmidhuber97} across the entire probability distribution of the variable of interest. Thus, the strength of this approach in the context of our specific empirical problem is threefold: its ability to account for time dependency in the data, its flexibility in accommodating  for non-linearities in the relationships between explanatory variables and the spread, and the possibility of focusing on specific portions of the distribution of the target variable. 
We calculate the forecast losses associated with 10 equally spaced quantiles of the probability distribution of the time series forecasts augmented with news and compare them to those associated to a baseline model containing only traditional determinants of government yield spreads. We also evaluate the usefulness of the DeepAR model in our empirical application by comparing the forecasting performance of our news-augmented model using DeepAR with that of the Gradient Boosting (GB) approach \citep{Hastie2009,Natekin2013GBM}, a trees based method that has shown to perform well relative to neural network techniques in predicting bond returns \citep{Bianchi_Buchner_Tamoni_2020}. The aim of this comparison is to disentangle the improvement in the forecasting power due to the inclusion of news indicators versus that linked to our deep learning approach. Our results show that news-based indicators extracted from GDELT used in combination with DeepAR greatly improve the forecasting performance across all quantiles of the distribution of the 10-year Italian government yield spread. In addition we find that using the DeepAR approach seems particularly important for quantiles belonging to the central part of the distribution, while it seems less important for extreme quantiles.

The rest of the paper is structured as follows. Section \ref{related} reviews existing literature on this topic, Section \ref{data} describes the data, while Section \ref{Model} illustrates the employed methodology. 
Section \ref{experiments} discusses the results of the analysis.
Finally, Section \ref{conclusions} concludes and highlights possible directions of future research.


\section{Background literature: time series forecasting with multi-layered neural networks} \label{related} 

In the last decades, a large number of studies have been adopting advanced machine learning methods to forecast economic and financial indexes \citep{Cavalcante2016194,Henrique2019226}. 
Recently, a vast literature has investigated the use of Deep Neural Networks \citep{Lecun2015436,SchmidhuberNeuralNetworks} for time series forecasting \citep{NeuralForecasting,Januschowski2020167}. Deep learning algorithms use a huge amount of unsupervised data to automatically extract complex representation. Relative to conventional machine learning techniques, such as support vector machines, gradient boosting or random forests, deep learning offers the advantage of unsupervised feature learning, strong generalization capabilities, and a robust training power for big data \citep{Lecun2015436,Huang2020}. Given the exponential implementation of deep learning in many sectors, including, among others, computer vision, healthcare, and audio-visual recognition \citep{Lecun2015436,Thaler201975,SchmidhuberNeuralNetworks}, it has recently attracted great attention in the field of economic and financial forecasting as well \citep{Huang2020}.\\~\

The use of Neural Networks (NNs) for time series forecasting is well surveyed in \citep{ijf1998}. Recent advances in the asset pricing literature \citep{Gu_Kelly_Xiu_2020,Bianchi_Buchner_Tamoni_2020}
show the ability of this approach to improve stock and bond returns predictability relative to classical regression-based methods.
When adopting a NN approach for time series forecasting, one possibility is to use Multilayer Perceptrons and address the sequential nature of the data by treating time as an explicit part of the input. However, such an approach has a number of important drawbacks, including the inability to process sequences of varying length and detect time-invariant patterns in the data.\\~\ 

A more suitable alternative consists in the use of Recurrent Neural Networks (RNNs) \citep{Graves2009545RNNs,Lecun2015436}. Under this approach, recurrent connections linking the neural networks' hidden units back to themselves with a time delay, allow handling the sequential nature of the data \citep{Deng20201543}. Examples of successful applications of RNNs in finance include forecasting of volatility and stock market movements \citep{Kraus201738,LienMinh201855392,Singh201718569,Jiang2018}, stock trading \citep{Deng2017653,Jeong2019125}, and portfolio management \citep{Almahdi2017267}.\\~\

%
Although RNNs are widely used in many practical financial forecasting applications, it turns out that training them is quite difficult given that they are typically applied to very long sequences of data \citep{Bao2017}. A common issue when training deep neural networks by gradient-based methods using back-propagation is that of vanishing or exploding gradients, which renders learning impossible. To address this issue, \citet{Schmidhuber97} proposed the so-called 
Long Short-Term Memory Networks (LSTMs). 
Instead of using a simple network at each step, LSTMs consider a more complex architecture composed of a cell and gates which manage input flows. Gates decide what information should be retained inside the cell and what should be propagated to the next time step \citep{Sutskever20143104}.
The cell has a memory state that is propagated across time along with the output of the LSTM unit, which is itself a function of the cell state. 
Unlike the output of the LSTM unit, the cell state undergoes minimal changes across time, thus the derivative with respect to the cell state does not decay or grow exponentially \citep{Gers2001669}. Consequently, there is at least one path where the gradient does not vanish or explode making LSTMs suitable for processing long sequences \citep{Sutskever20143104}. 
LSTMs have seen huge success in a wide range of financial forecasting applications such as stock market prediction \citep{Sohangir2018,Kim201825,Baek2018457}, stock trading \citep{Bao2017,Fischer2018654}, banking default
risk and credit \citep{Wang20192161,Jurgovsky2018234}, among many others.\\~\

Convolutional Neural Networks (CNNs), forming the other most popular deep learning models family, have been also applied to the field of financial time series forecasting \citep{Chen201787}. CNNs, however, are much more suitable at classifying images, and, generally, high dimensional structures, rather than time series \citep{Lecun2015436}.
CNNs are best known indeed as high performing image classifiers \citep{Ciresan20111237}.
Nevertheless, some applications of CNNs to financial time series forecasting do exist in the literature, e.g. for the prediction of the stock market \citep{Vargas201760} and changes in foreign exchange rates \citep{Galeshchuk2017100}. Another proven CNNs approach for financial forecasting is the Dilated Convolutional Neural Network presented in \citep{Borovykh2017729}, wherein the underlying architecture comes from DeepMind’s WaveNet project \citep{wavenetArxiv}.
Very recently, \citet{Salinas2020} have proposed an RNN-based forecasting model using LSTM or Gated Recurrent Units (GRUs) \citep{Schmidhuber97}, 
the latter being a simplification of LSTMs that do not use a separate memory cell and may result in a good performance for certain applications. 
At each time step this approach, also known as DeepAR, takes as input the previous time points and covariates, and estimates the distribution of
the value for the next period. This is done via the estimation of the parameters
of a pre-selected parametric distribution (e.g. negative binomial, student $t$, Gaussian, etc). Training and prediction follow the general approach for auto-regressive models \citep{Salinas2020}. One feature that makes this setting appealing is that, rather producing a single forecast, it allows to carry inference on the entire predictive distribution, thus 
making the analysis more robust and reducing uncertainty in the downstream decision-making flow. In our empirical study we adopt this method.




\section{Data} \label{data} 
\subsection{Financial data} \label{infrastracture} 

\begin{figure}[t!]
\centering
\begin{subfigure}[b]{0.79\textwidth}
 \centering
 \includegraphics[width=\textwidth]{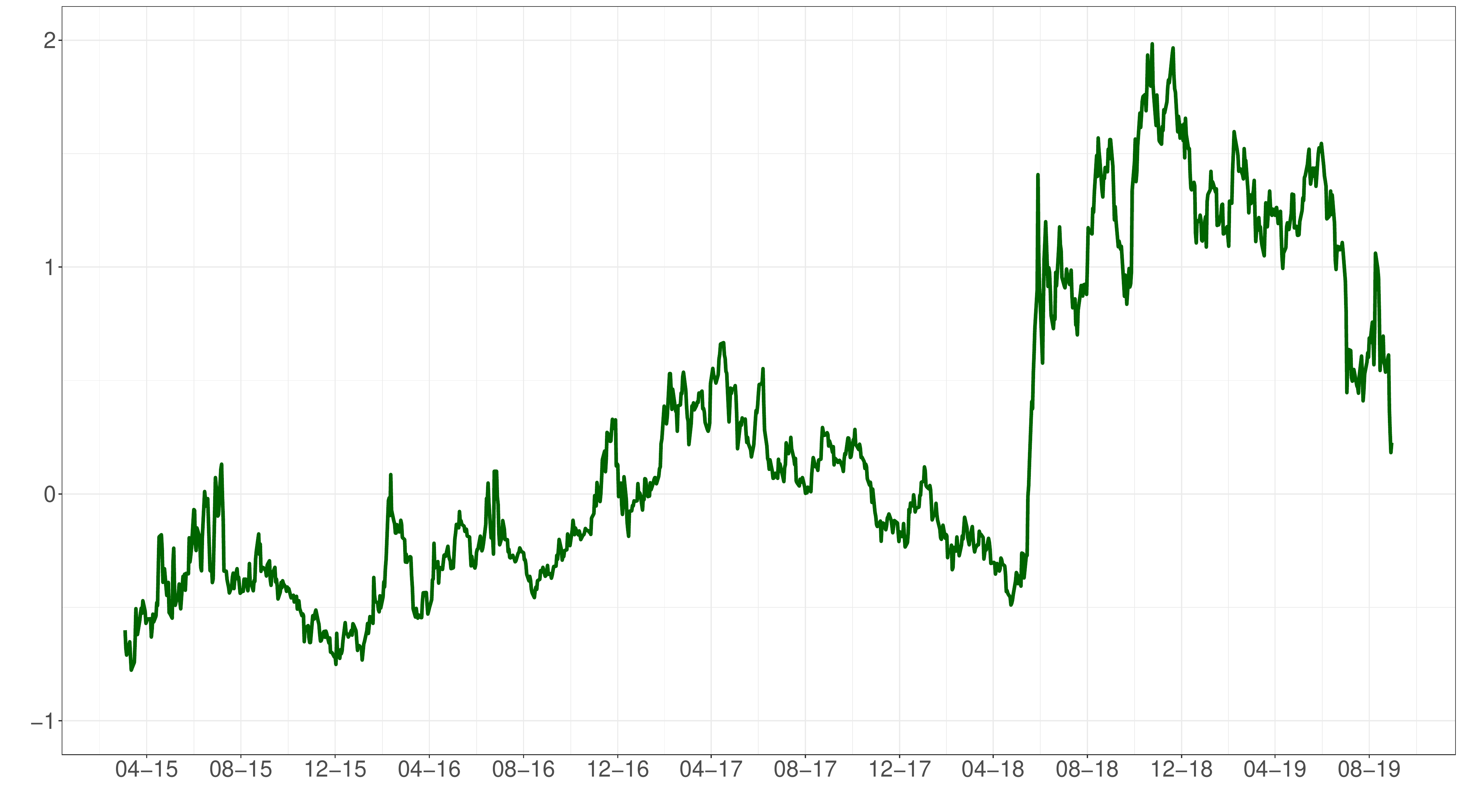}
\end{subfigure}
\vfill
\bigskip
\begin{subfigure}[b]{0.79\textwidth}
 \centering
 \includegraphics[width=\textwidth]{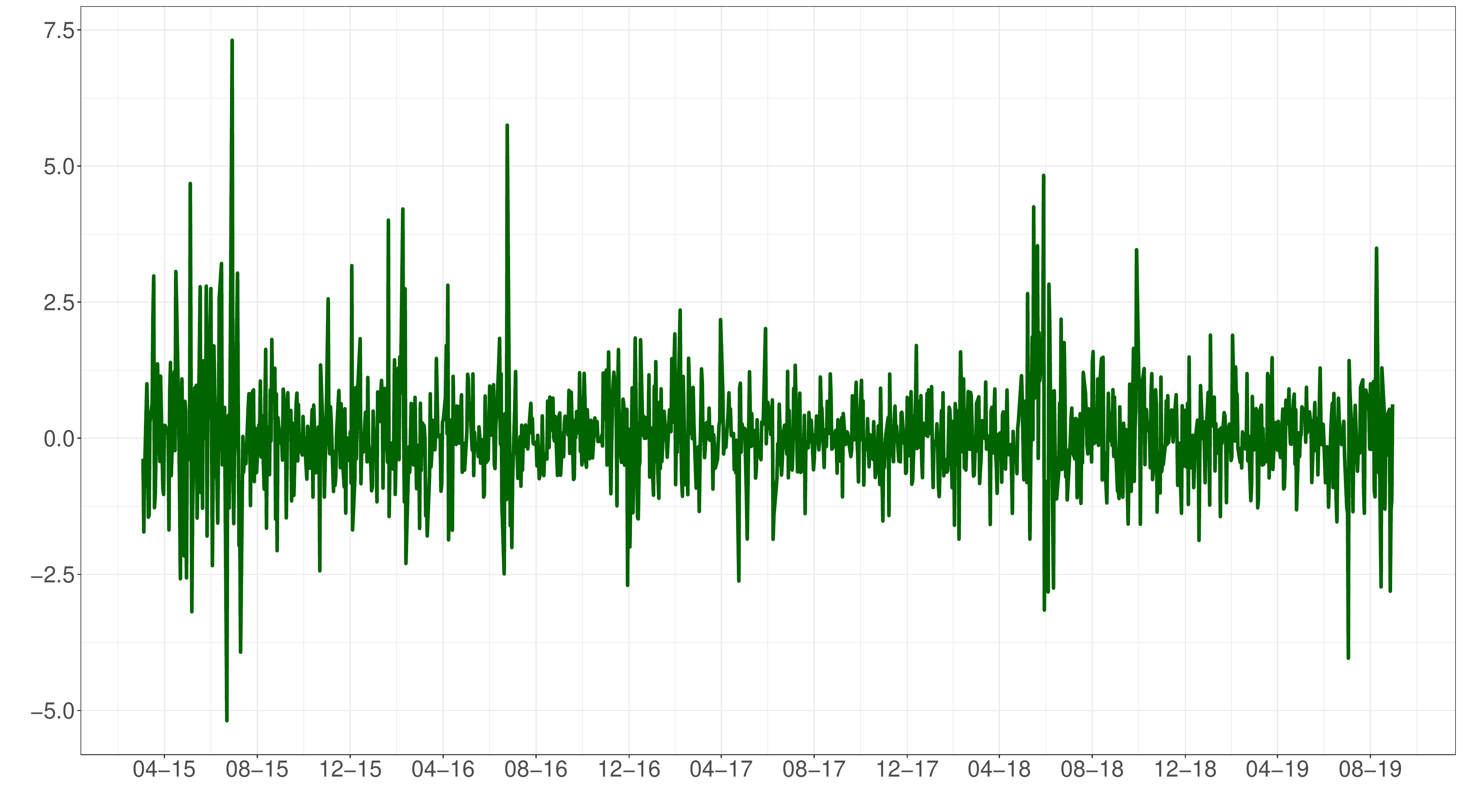}
\end{subfigure}
\vfill
\caption{Sovereign spread for Italy against Germany as the difference between the Italian 10 year maturity bond yield minus the German counterpart in level (top) and in log-differences (bottom). Data have been extracted from Bloomberg.}
\label{target}
\end{figure}

We have extracted daily government interest rates from Bloomberg\footnote{\url{https://www.bloomberg.com/professional/product/market-data/}}, for Italy and Germany over the period from the 2nd of March 2015 to the 31st of August 2019. We have considered all maturities from 3 months up to 10 years. Missing data relative to weekends have been dropped 
from the target time series, yielding a total of $1,174$ data points.
Using these data, we have calculated the sovereign spreads at maturity $\tau$, as the difference between the Italian and German interest rate yields on maturity $\tau$. Our empirical exercise will focus on forecasting the sovereign spread at maturity 10 years, namely, $Spread_{t}(10)$. The time evolution of this series is represented in the upper graph of Figure \ref{target}. This graph shows a striking increase in the target variable in May 2018, when uncertainty about the formation of a stable government in Italy was a major concern amongst investors. The Italian spread remained at high levels during that summer while it increased further in October and November 2018, when investors worried about the possibility of Italian deficit spending engagements breaking European fiscal rules.

As the government bond yields are highly persistent and non-stationary, as also evident from its temporal pattern displayed in the top graph of Figure \ref{target}, we have applied the log-difference transformation to obtain a stationary series of daily changes. In addition, to attenuate the effect of outliers, we have applied a robust scaling transformation to the resulting series by subtracting its median and dividing by its interquartile range. The temporal evolution of the resulting variable, which is our target variable, is displayed at bottom of Figure \ref{target}. Finally, we have extracted from our data the so-called level, slope and curvature factors of the term structure of interest rates spreads, also known as \textit{yield spread curve}. Advanced by \citet{Nelson_Siegel_87}, these factors are considered by the existing literature in finance as the most relevant piece of information used by investors to form their expectations about future rates dynamics \citep{Sargent_72}. Several empirical papers have shown that these three factors, also known as Nelson and Siegel term-structure factors, account for most of the variability in interest rates \citep{D_S_03,Litterman_Scheinkman_91}. We estimate the factors by following the procedure outlined by \citet{Diebold_Li_06}. 
Specifically, for each day we regress the entire government yield spread curve on a constant and two factor loadings $\left( \frac{1 - e^{-\lambda_t \tau}}{\lambda_t \tau} \right)$ and $\left( \frac{1 - e^{-\lambda_t \tau}}{\lambda_t \tau} - e^{-\lambda_t \tau} \right)$, where $\lambda_t$ is a constant affecting the hump shape of curvature factor that is assumed to be equal to 0.0609. The time-series of the estimated parameters $\hat{\beta}_{0,t}$, $\hat{\beta}_{1,t}$, $\hat{\beta}_{2,t}$ are the factors that will be used as regressors in our forecasting models in addition to our news indicators.

\subsection{News data}


GDELT is an open, big data platform of meta-information extracted from broadcast, print, and web news collected at worldwide level and translated nearly in real-time into English from over 65 different languages \citep{Leetaru_2013gDELT}. 
Three primary data streams are created, one codifying human activities around the world in over 300 categories, one recording people, places, organizations, millions of themes and thousands of emotions underlying those events and their interconnection, and one codifying the visual narratives of the world's news imagery. 
The output of such processing is updated on the GDELT website and is freely available to users by means of custom REST APIs. 
Extracted and processed information are stored into different databases, with the most comprehensive among these being the GDELT Global Knowledge Graph (GKG). GKG is a news-level data set, containing a rich and diverse array of information. Specifically, for each news it provides information on people, locations and organizations mentioned in the article and retrieves counts, quotes, images and themes using a number of popular topical taxonomies, such as the World Bank Topical Ontology \citep{WorldBankTopicalTaxonomy}, or the GDELT built-in topical taxonomy \citep{Leetaru_2013gDELT}. 
GDELT also measures a large number of emotional dimensions expressed by means of a number of popular dictionaries in the literature. In total there are 51 dictionaries, including 
the ``Harvard IV-4 Psychosocial Dictionary'' \citep{stone1966general}, 
the ``WordNet-Affect dictionary'' \citep{Strapparava2004,Strapparava2004a}, 
the ``Loughran and McDonald Sentiment Word Lists dictionary'' \citep{loughran2016textual,Loughran_McDonald_2011}, 
among others. Given that each dictionary carries a large number of dimensions, we can observe a total of 2,230 dimensions, known as Global Content Analysis Measures (GCAMs) \citep{Leetaru_2013gDELT,Kwak2014}. 
In terms of volume, GKG analyses over 88 million articles a year and more than 150,000 news outlets. Its dimension is around 8 TB, growing approximately 2TB each year. 
Until now GDELT has been used in the analysis of political conflicts \citep{Hammond2014}, for the monitoring of 
diseases and natural disasters around the world \citep{Kwak2014} and recently it has been employed to study the evolution of Sustainable Development Goals (SDG) at global scale \citep{Czvetko_Honti_Sebestyen_Abonyi_2021}. However, its use for investigating the impact of news on financial markets and macroeconomic forecasting, is limited to a few studies, such as those by \citet{Acemoglu2017} and \citet{Tilly_Ebnerb_Livanc_2021}. 

For our application, we use the GKG fields from the World Bank Topical Ontology, 
the emotional dimensions available through the GCAMs, and the name of the journal outlets.
The huge amount of unstructured documents coming from GDELT has been re-engineered and stored into an ad-hoc Elasticsearch infrastructure \citep{elastic2015,Shah2018}. Elasticsearch is a popular and efficient document-store built on the Apache Lucene search library, providing real-time search and analytics for different types of complex data structures, like text, numerical data, or geospatial data, that have been serialized as JSON documents. Elasticsearch can efficiently store and index data in a way that supports fast searches, allowing data retrieval and aggregate information functionalities via simple REST APIs to discover trends and patterns in the stored data. 
Without Elasticsearch, we wouldn't have been able to interact and search over a so large amount of textual, unstructured data.
 
We have extracted news information from GKG from a set of around 20 Italian newspapers (see Appendix A), published over the period 2 March 2015 to 31 August 2019. 
We rely on the Geographic Source Lookup file available on the GDELT platform \citep{Leetaru_2013gDELT,Kwak2014} in order to chose both generalist national newspapers with the widest circulation in Italy, as well as specialized financial and economic outlets. To extract news whose main themes are related to events concerning bond market investors, we have exploited information from the World Bank Topical Ontology (WB) to understand the primary focus (theme) of each article. Such taxonomy is a classification schema for describing the World Bank’s areas of expertise and knowledge domains representing the terminologies and conceptual entities used by domain experts and World Bank workforce. Hence, we select only articles such that the topics extracted by GDELT fall into one of the following WB themes of interest: \textit{Macroeconomic Vulnerability and Debt}, and \textit{Macroeconomic and Structural Policies}. We observe that articles can mention only briefly one of the selected topics and then focus on a totally different theme. To make sure that the main focus of the article is one of the selected WB topics, we retain only news that contain in their text more than three keywords belonging to these themes. The aim is to select news that focus on topics relevant to the bond market, while excluding stories that only briefly report macroeconomic, debt and structural policies issues. Finally, to obtain a pool of news that are not too heterogeneous in length, we have considered only articles that are at least 100 words long. After this selection procedure we are left with a total of 21,313 articles over a period of 5 years length.

Once collected the relevant news data, we have mapped this information to the relevant trading day. Specifically, we assign to a given trading day all the articles published during the opening hours of the bond market, namely between 9.00 am and 17.30 pm. Furthermore, articles that are published after the closure of the bond market or overnight are assigned to the following trading day. Following \citet{Garcia2013}, we assign the news published during weekends to Monday trading days. Finally, since the GKG operates on the UTC time, 
we make a one-hour lag adjustment according to the Italian time zone (i.e. UTC+1).  

\subsection{Feature selection} \label{features} 
We have aggregated GDELT data at the level of the trading day by counting the total number of words belonging to each category within each day, where our categories are GCAMs, themes, persons, organisations, and locations. By doing this, we obtain a data set of 10,489 features, of which 2,978 are GCAMs, 2,106 are themes, 871 are persons, 4,374 are organisations, and 160 are locations, observed over 1,174 trading days. 

From such a large amount of variables, we have applied a number of 
criteria as well as domain knowledge to extract a set of pertinent features and discard inappropriate and redundant variables.
First, we have dropped from the analysis all GCAMs for non-English language and those that are not relevant for our empirical context (for example, the Body Boundary Dictionary), thus reducing the number of GCAMs to 407 and the total number of features to 7,916. We have then discarded variables with an excessive number of missing values within the sample period. In particular, we have dropped variables that contain all missing values in the first 33 percent of the sample period, given that this part of the sample is used for estimation purposes. Of the remaining features, we have only retained those that are available at least 90 percent of the total number of days. Finally, we have excluded variables with very low variability by keeping only those that have a standard deviation, calculated over the full sample, greater than 5 words.
By applying these filters, the features in the data set are reduced to 17 themes, 149 GCAMs and 2 locations, while features related to persons and organisations are completely dropped out from the study.
We finally perform a correlation analysis across the selected variables, after having normalised them by dividing each feature by the number of daily articles. If the correlation between any two features is above 70$\%$, we give preference to the variable with an inferior number of missing values, while, in the case of ties, we consider the following order of priority among categories: GCAMs, WB themes, GDELT themes, and locations. In the rare event of an identical number of missing values and the same category of two highly correlated variables, we simply pick one of them at random. After this feature engineering procedure, we are left with a total of 51 GDELT variables, of which 11 are themes, 39 are GCAMs, and 1 is a location (i.e., Germany). The final set of variables is composed of 51 selected features listed in Appendix B.

\begin{figure}[!ht]
\centering
\begin{subfigure}[b]{0.49\linewidth}
 \centering
 \includegraphics[width=\textwidth]{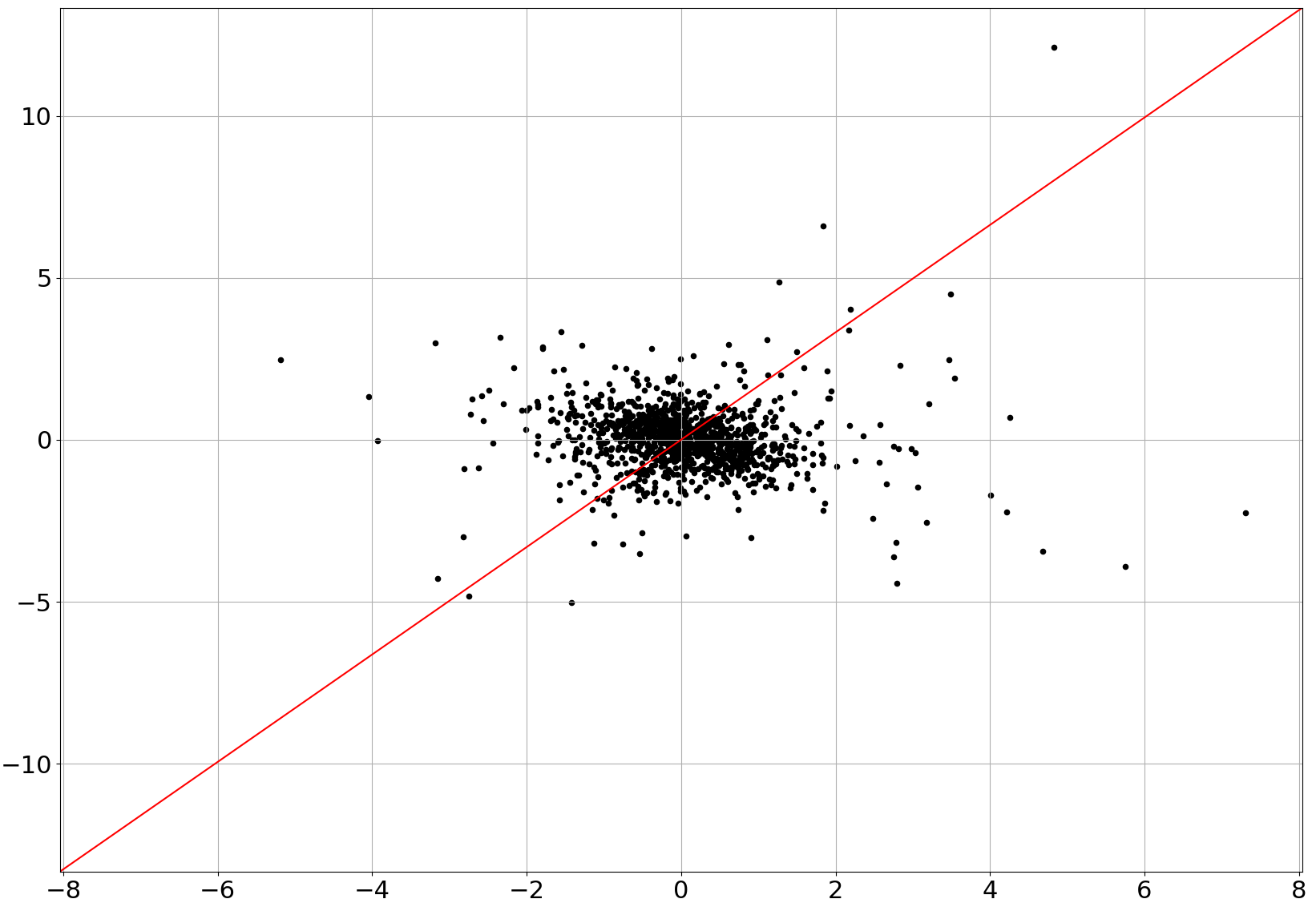}
 \caption{\small{\emph{FACTOR\_1}.}}
\end{subfigure}
\hfill
 \medskip
\begin{subfigure}[b]{0.49\linewidth}
 \centering
 \includegraphics[width=\textwidth]{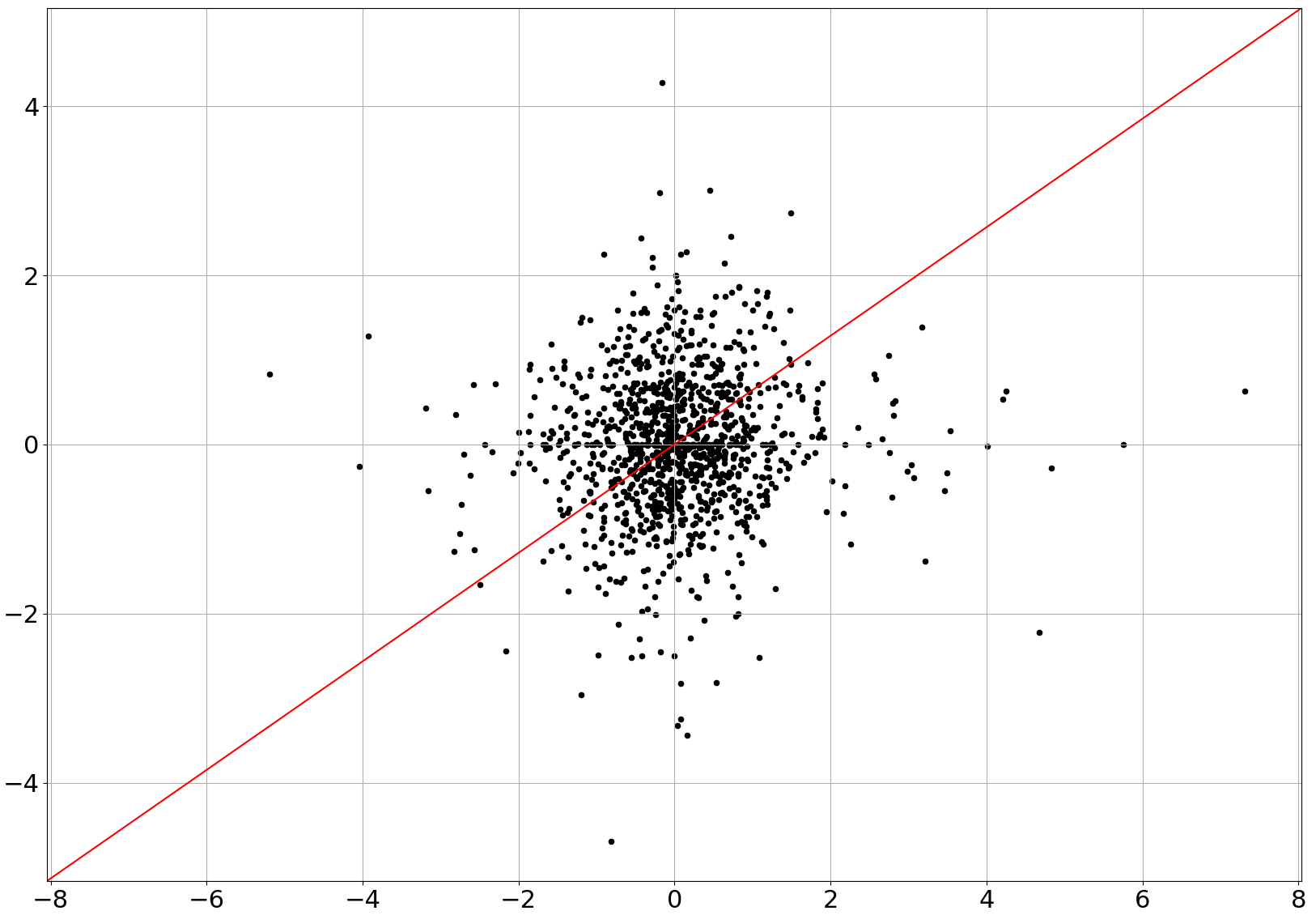}
 \caption{\small{\emph{GCAM c9.297}.}}
\end{subfigure}
\hfill
 \medskip
\begin{subfigure}[b]{0.49\linewidth}
 \centering
 \includegraphics[width=\textwidth]{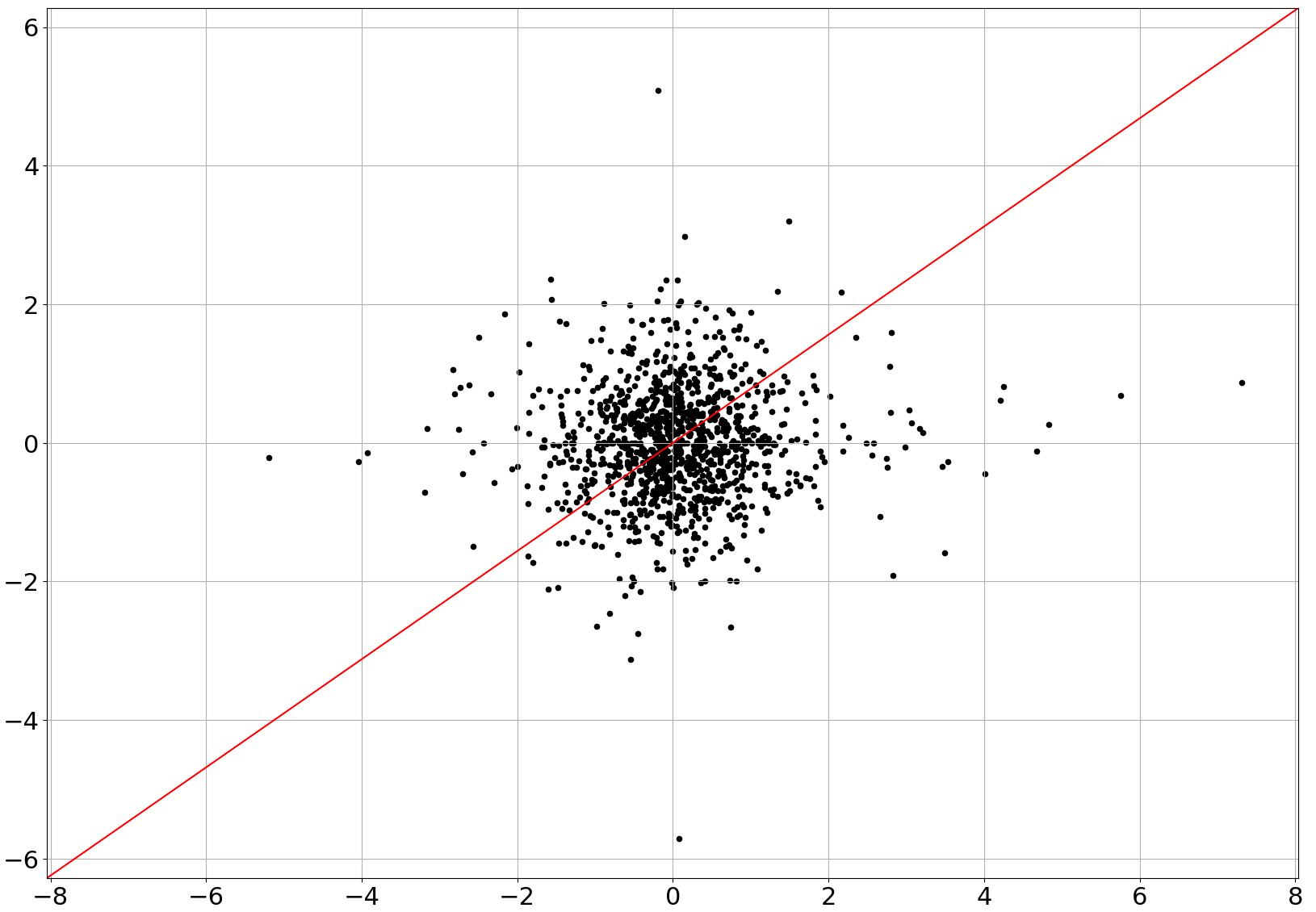}
 \caption{\small{\emph{GCAM c2.168}.}}
\end{subfigure}
\hfill
 \medskip
\begin{subfigure}[b]{0.49\linewidth}
 \centering
 \includegraphics[width=\textwidth]{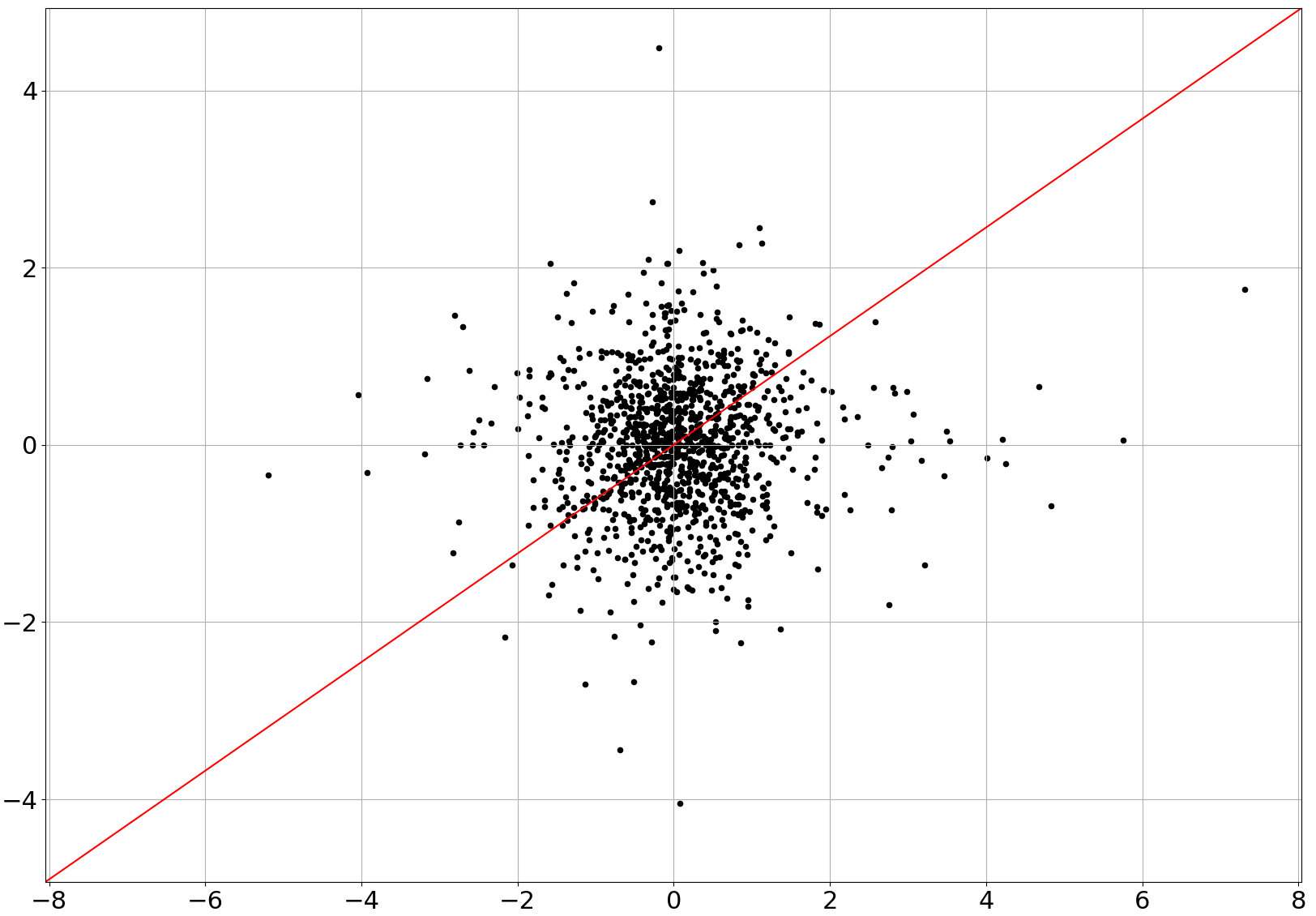}
 \caption{\small{\emph{GCAM c15.86}.}}
\end{subfigure}
\hfill
\caption{
Scatter plots 
of the sovereign spread for Italy against Germany versus its top correlated covariates that are, respectively: (a) the first Nelson and Siegel term-structure factor (FACTOR\_1), (b) the Roget's Thesaurus GCAM ``Recession'' (c9.297), (c) the Harvard IV-4 GCAM ``Loss'' (c2.168), and (d) the WordNet Affect GCAM ``Distress'' (c15.86).}
\label{correlations-scatter}
\end{figure}

It is interesting to observe that the selected topics contain WB themes such as Inflation, Monetary Policy, Central Banks, Debt and GDELT themes such as Government and Taxation that are indeed important topics discussed in the news when considering interest rates issues. Instead, the selected GCAMs include features like optimism, pessimism or arousal, that support expressing indeed the emotional state of the market. 
Figure \ref{correlations-scatter} reports the scatter plots of the target variable versus
the four most highly correlated covariates, where we see that, as expected, the most highly correlated feature is the first Nelson and Siegel term-structure factor, i.e. \emph{FACTOR\_1}. It is interesting to note also that the 
other highly correlated features are, respectively, the Roget's Thesaurus GCAM ``Recession'' (c9.297), the Harvard IV-4 GCAM ``Loss'' (c2.168), and the WordNet Affect GCAM ``Distress'' (c15.86), which are variables strongly connected also intuitively to periods of high sovereign spread volatility.


\section{Methods} \label{Model}

\subsection{Neural forecasting with DeepAR} \label{deepar-meth}

Artificial neural networks \citep{RipleyNeuralNetworks,ijf1998} are popular machine learning approaches which mimic the human brain and represent the backbone of deep learning algorithms \citep{SchmidhuberNeuralNetworks}. %
A neural network is based on a collection of connected units or nodes, called artificial neurons, which loosely model the neurons in a biological brain. Each connection, like the synapses in a biological brain, can transmit a signal to other neurons. An artificial neuron receives and processes a signal, and then transmits the output to other neurons connected to it. The ``signal'' at a connection is a real number, and the output of each neuron is computed by some non-linear function of the sum of its inputs. Neurons and their connections typically have a weight that adjusts as the learning process proceeds. The weight increases or decreases the strength of the signal at a connection. Neurons may have a threshold such that a signal is sent only if the aggregate output crosses that threshold \citep{RipleyNeuralNetworks}. Typically, neurons are aggregated into layers, which may perform different transformations on their inputs. It is indeed the number of node layers, or the depth, of neural networks that distinguishes a single artificial neural network from a deep learning algorithm, which must have more than three \citep{SchmidhuberNeuralNetworks}. Signals travel from the first layer (the input layer), to the last layer (the output layer), possibly after traversing the layers multiple times.

Recurrent Neural Networks (RNNs) are a special class of artificial neural networks where connections between nodes form a directed or an undirected graph along a temporal sequence \citep{Graves2009545RNNs,Lecun2015436}. This allows them to exhibit temporal dynamic behaviour, and to 
use their internal state (memory) to process variable length sequences of inputs. 
%
%
When processing input sequences, RNNs apply a recurrence at each time step \citep{Graves2009545RNNs}. 
Furthermore, model size does not increase with the size of the input, which
theoretically can be any length. RNNs allow layers to have stored hidden states when being input to the next layer, so that historical values can play a role in prediction (although accessing very early data in an input sequence can be relatively difficult). The storage of hidden states can be under direct control by the neural network, and might be also replaced by another network or graph, if that incorporates time delays or has feedback loops \citep{Lecun2015436,Graves2009545RNNs}. Such controlled states are referred to as gated state or gated memory, and are part of Long Short-Term Memory Networks (LSTMs) and Gated Recurrent Units (GRUs) \citep{Schmidhuber97}. 

Proposed by \citet{Salinas2020}, DeepAR is a neural forecasting method based on RNNs that leverages previous work on deep learning methods developed specifically for forecasting time series data \citep{Lecun2015436,NeuralForecasting,Januschowski2020167}. 
In essence, DeepAR tailors an LSTM-based architecture 
into a probabilistic forecasting setting, that is, predictions are not restricted to point forecasts only, but probabilistic forecasts are produced accordingly to a user-defined predictive distribution \citep{NeuralForecasting,Januschowski2020167}. 
The outcome is more robust relative to point forecasts, and uncertainty in the downstream decision-making flow is reduced by minimising expectations of the adopted negative log-likelihood loss function under the forecast distribution \citep{Salinas2020}.

LSTMs have been originally proposed to solve the so-called vanishing or exploding gradient problem, typical of RNNs \citep{Schmidhuber97}. These problems arise during back-propagation in the training of a deep network, when the gradients are being propagated back in time all the way to the initial layer \citep{Greff20172222}. The gradients coming from the deeper layers have to go through continuous matrix multiplications because of the chain rule. As they approach the earlier layers, if they have small values ($<<1$), they shrink exponentially until they vanish, thus making it impossible for the model to learn any further. This is known in the literature as the vanishing gradient problem. 
On the other hand, the opposite situation, known as the exploding gradient problem, arises when the gradients have large values ($>>1$). In this case they get larger and larger and eventually blow up. 
To solve these issues, LSTMs employ computational gates to control information flow over time and usually perform better with long-term dependencies \citep{Sutskever20143104}.


\begin{algorithm}[t!]
\SetAlgoVlined
\KwIn{ $x = [x_1,\cdot, x_n]$, $x_t \in \mathbb{R}$ } 
\KwOut{ $h = [h_1,\cdot, h_n]$, $h_t \in \mathbb{R}$ }
\emph{Initialization:}\\
- Let $x_t$, $C_t$ and $h_t$ be, respectively, the input, cell state, and output of a cell $t$;\\
- Let $[w_f, w_i, w_C, w_o]$ and $[b_f, b_i, b_C, b_o]$ be learned weights and bias coefficients in $\mathbb{R}$;\\
- Let g and tanh be the sigmoid and tanh activation functions;\\
\ForEach{$x_t \in x$}{
$\cdot$~Calculate $f_t \leftarrow$\emph{forget-gate($g,w_f,b_f,h_{t-1},x_t$)}\;
$\cdot$~Calculate $i_t \leftarrow$\emph{input-gate($g,w_i,b_i,h_{t-1},x_t$)}\;
$\cdot$~Calculate $\hat{C}_t \leftarrow$\emph{cell-state-gate($tanh,w_i,b_i,h_{t-1},x_t$)}\;
$\cdot$~Update the cell state: $C_t=f_t \cdot C_{t-1} + i_t \cdot \hat{C}_t$\;
$\cdot$~Calculate $o_t \leftarrow$\emph{output-gate($g,w_o,b_o,h_{t-1},x_t$)}\;
$\cdot$~Update the output: $h_t = o_t \cdot tanh(C_t)$\;
}
$\Rightarrow$~\emph{Return($h$)}. 
\caption{Pseudocode of LSTM layer} \label{LSTM-alg}
\end{algorithm}

An LSTM cell, say $t$, having input $x_t$, cell state $C_t$, and output $h_t$, is composed by four different interacting gate layers \citep{Greff20172222}, namely \emph{forget}, \emph{input}, \emph{cell state} and \emph{output} layers, described in pseudocode in Algorithm \ref{LSTM-alg}.
%

The ``forget gate layer'' decides the amount of information from the past cell state $t-1$ to forget or to keep in the current cell $t$.
This is done by looking at its input $x_t$ and at the previous cell output $h_{t-1}$, and for each number in the cell state $C_{t-1}$ a value, $f_t$, between 0 and 1 is produced by the following formula: $f_t=g(w_f\cdot[h_{t-1},x_t]+b_f)$, where $w_f$ are weights associated to the forget gate $f$, $b_f$ are its bias coefficients, and $g$ is the activation function, which can be, for instance, a ReLU, a tanh or a sigmoid function \citep{Sutskever20143104}. 
For the outcome $f_t$, the value $1$ means ``completely keep this'' while the value $0$ represents ``completely get rid of this''.

Instead, the ``input gate layer'' decides the amount of each previous cell state values to update with respect to the input $x_t$ and the previous cell output $h_{t-1}$. This is done by computing another value, say $i_t$, by means of an activation function $g$ (e.g., a sigmoid): $i_t=g(w_i\cdot[h_{t-1},x_t]+b_i)$, where $w_i$ are the weights associated to the input gate $i$, and $b_i$ are its bias coefficients.

The ``cell state gate layer'' decides the portion of new information that is going to be stored in the cell state $C_t$. This is done by a tanh activation function that, by looking at the input $x_t$ and at the previous cell output $h_{t-1}$, creates a vector of new candidate values, say $\hat{C}_t$, that could be added to the state: $\hat{C}_t=tanh(w_C\cdot[h_{t-1},x_t]+b_C)$, with $w_C$ and $b_C$ being, respectively, the weight coefficients associated to the cell state gate and its bias coefficients. 
After that, the new cell state $C_t$ is obtained by the previous cell state $C_{t-1}$ by putting together the already computed pieces into the following formula: $C_t=f_t \cdot C_{t-1} + i_t \cdot \hat{C}_t$.
In words, the old state $C_{t-1}$ is multiplied by the forget coefficient $f_t$, and the result is added to the new candidate values $\hat{C}_t$ scaled by $i_t$ (i.e., the quantity of each state value that has been decided to update).

Finally, given the computed cell state $C_t$, it is possible to compute the output $h_t$ of the LSTM cell by means of the ``output gate layer''. 
First, it is computed the amount, say $o_t$, of the cell state that will be propagated by the following formula: $o_t=g(w_o\cdot[h_{t-1},x_t]+b_o)$, where $w_o$ are the weight coefficients associated to the output gate, $b_o$ are its bias values, and $g$ is the activation function, which in our case is a sigmoid. 
Then, we tanh-transform the cell state (to force the values to be between -1 and 1) and multiply it by the previous output $o_t$, so that we only output the parts that are needed: $h_t = o_t \cdot tanh(C_t)$. 
The output $h_t$ of the current cell, $t$, is obtained, and it is then propagated to the next LSTM cell, $t+1$, along with its state $C_t$, where the whole four layers mechanism is then iterated \citep{Sutskever20143104}.
All the weights and bias coefficients, respectively $w_f, w_i, w_C, w_o$ and $b_f, b_i, b_C, b_o$, are estimated during the training phase of the RNN by back-propagating and minimising the error (the negative log-likelihood in the case of DeepAR).

The DeepAR model implements such LSTM cells in an architecture that allows for simultaneous training of many related time-series and implements an encoder-decoder setup common in sequence-to-sequence models \citep{Salinas2020,Sutskever20143104}. In addition, DeepAR features a probabilistic forecasting setting that produces probability distributions for the forecasted variable. 
%
%
DeepAR is local in space and time, which means that the input length does not affect the storage requirements of the network. The computational complexity of learning each LSTM model per weight and time step is $\mathcal{O}(1)$ \citep{Schmidhuber97}.
Therefore, the overall 
computational complexity of a DeepAR model per time step is equal to $\mathcal{O}(w \cdot c)$, where $w$ is the total number of weights and $c$ the total number of LSTM cells.
The learning time for a DeepAR network with an input length $i$ and a number of epochs $e$ can be then calculated as $\mathcal{O}(w \cdot c \cdot i \cdot e)$. Therefore, we can conclude that our model has $\mathcal{O}$ complexity in the typical asymptotic notation.

In our application, we have implemented the DeepAR model developed with Gluon Time Series (GluonTS) \citep{gluonts-arxiv2019}, an open-source library 
for probabilistic time series modelling that focuses on deep learning-based approaches. 
We adopt a rolling window estimation technique for training and validation, with a window length equal to half of the full sample, that is $586$ data points. The first estimation sample, for example, starts at the beginning of March and ends in May 2017. For each window, we calculate one step-ahead forecasts. 
Hyperparameter tuning for the model \citep{Selvin20171643} has been performed through Bayesian hyperparameter optimization using the Ax Platform \citep{LethamBakshy2019,Bakshy2018AEA} on the first estimation sample, 
providing the following best configuration: 2 RNN layers, each having 40 LSTM cells, 500 training epochs, and a learning rate equal to 0.001, with training loss being the negative log-likelihood function.

In order to estimate the predictive power of GDELT features together with Nelson and Siegel term-structure factors we have considered a number of alternative specifications that include different sets of regressors. 
%
First, we have considered a pure autoregressive DeepAR model, that we refer to as \emph{DeepAR-NoCov}, which does not include any covariates in the model. We then consider a DeepAR model with the traditional Nelson and Siegel term-structure factors used as the only covariates, that we call \emph{DeepAR-Factors}. Next, we augment this model with the 51 pre-selected GDELT variables, yielding to the so-named \emph{DeepAR-Factors-GDELT} model.

\subsection{Variable reduction} \label{experiment-settings}

Given that our time series is of limited length, we also investigate whether a further data-driven feature reduction step could provide performance improvements. To this end, we have applied unsupervised hierarchical clustering \citep{Larose20051} to the three factors and GDELT variables together, and considered a varying number of clusters between 2 to 54. We have chosen as the optimal number of clusters \citep{Fraley1998586}, the configuration yielding the highest Silhouette width value \citep{Larose20051,Charrad20141} calculated by means of the Euclidean distances among the variables. The Silhouette width value is a measure of how similar an object is to one specific cluster compared to other clusters, and ranges between $-1$ and $+1$, with a high value indicating a high similarity versus small values pointing at a poor match. Once selected the clusters, the features with shorter Euclidean distance with respect to the cluster centroids are retained.
This procedure applied to the GDELT data jointly with the conventional interest rates factors has produced an optimal number of clusters equal to 24, having an average Silhouette score equal to 0.082. The set of retained features is composed by the three Nelson and Siegel term structure factors and 21 GDELT variables, indicated with a tick in the third column (i.e. \emph{Hierarc}) of Table \ref{GroupResQuantiles} in Appendix B.
The remaining features have been discarded and not further considered in the model. We refer to this model setting as \emph{DeepAR-Factors-GDELT-hierarc}.

\begin{figure}[!t]
\begin{center}
\begin{subfigure}{0.49\linewidth}
 \centering
 \includegraphics[width=\textwidth]{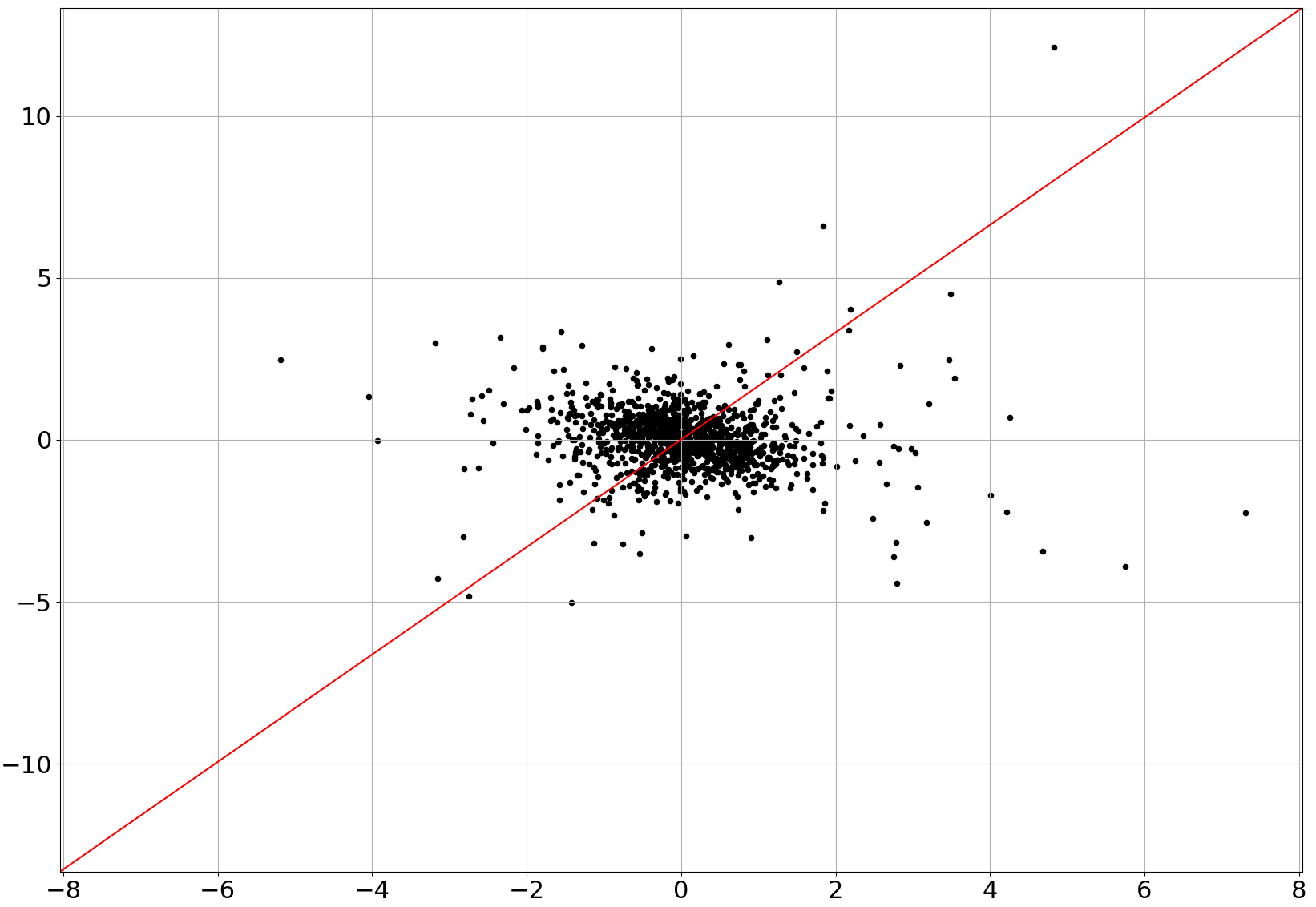}
 \caption{\small{\emph{FACTOR\_1}.}}
\end{subfigure}
\hfill
 \medskip
\begin{subfigure}{0.49\linewidth}
 \centering
 \includegraphics[width=\textwidth]{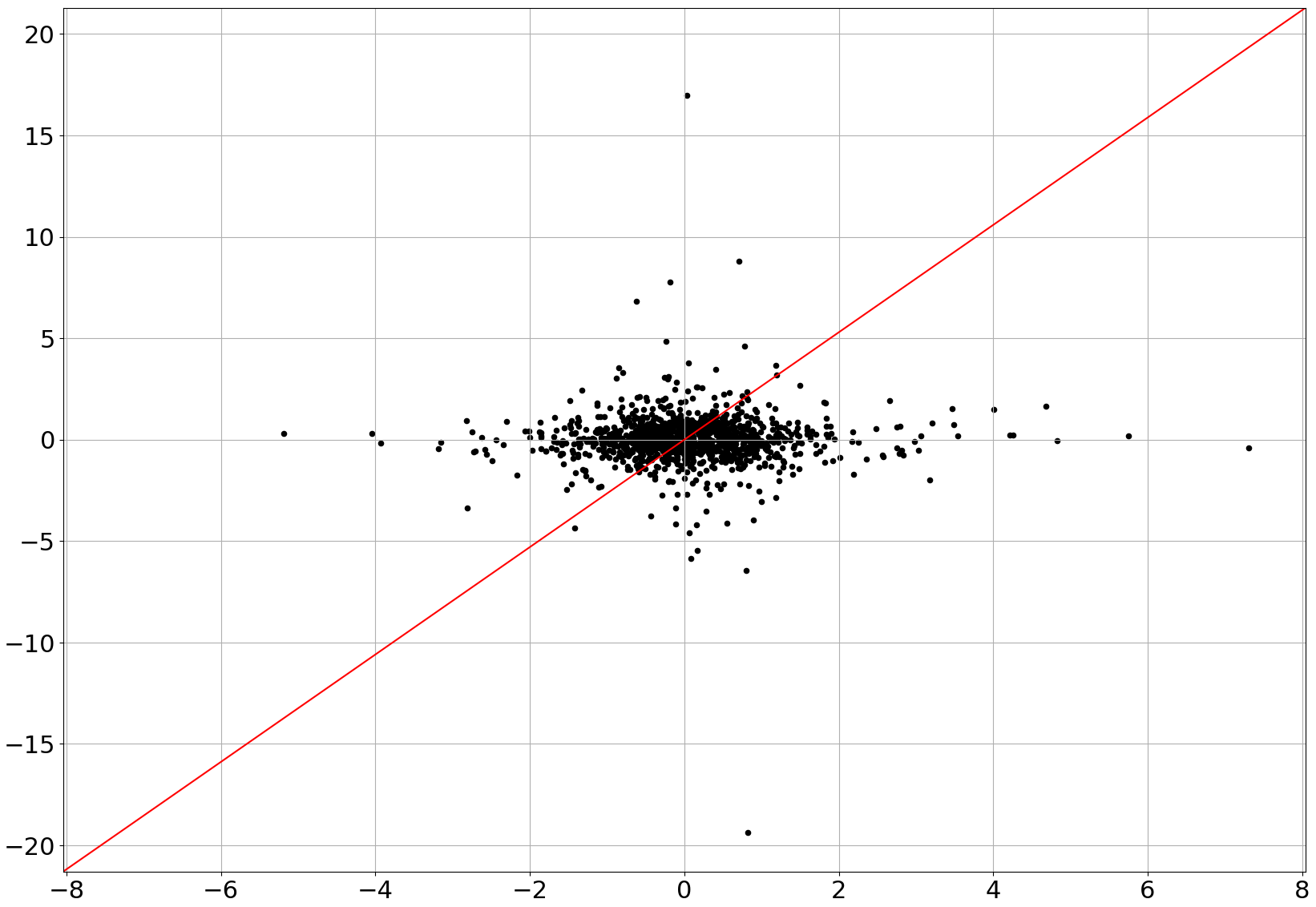}
 \caption{\small{\emph{PCA\_1}.}}
\end{subfigure}
\hfill
 \medskip
\begin{subfigure}{0.49\linewidth}
 \centering
 \includegraphics[width=\textwidth]{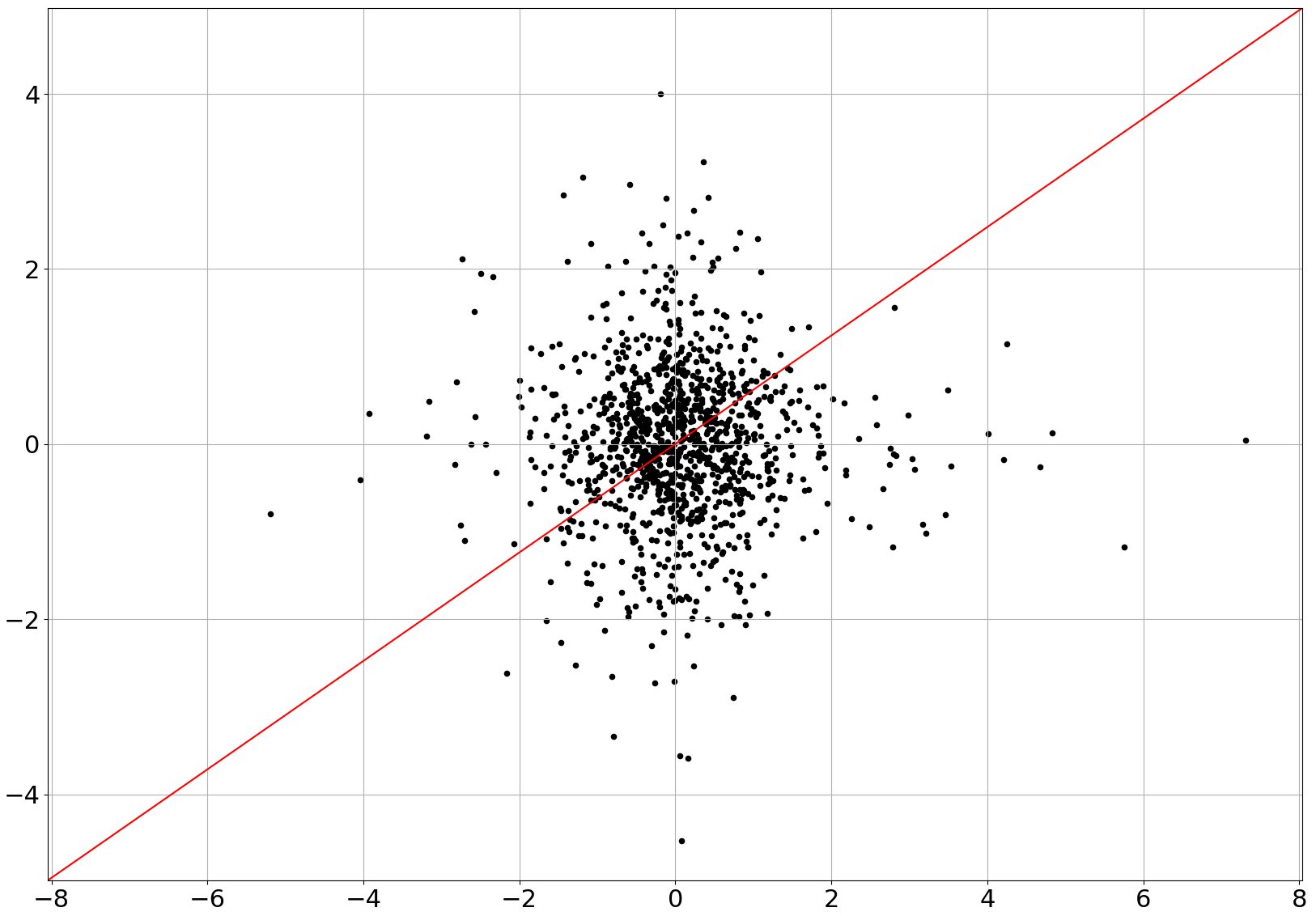}
 \caption{\small{\emph{PCA\_2}.}}
\end{subfigure}
\hfill
 \medskip
\begin{subfigure}{0.49\linewidth}
 \centering
 \includegraphics[width=\textwidth]{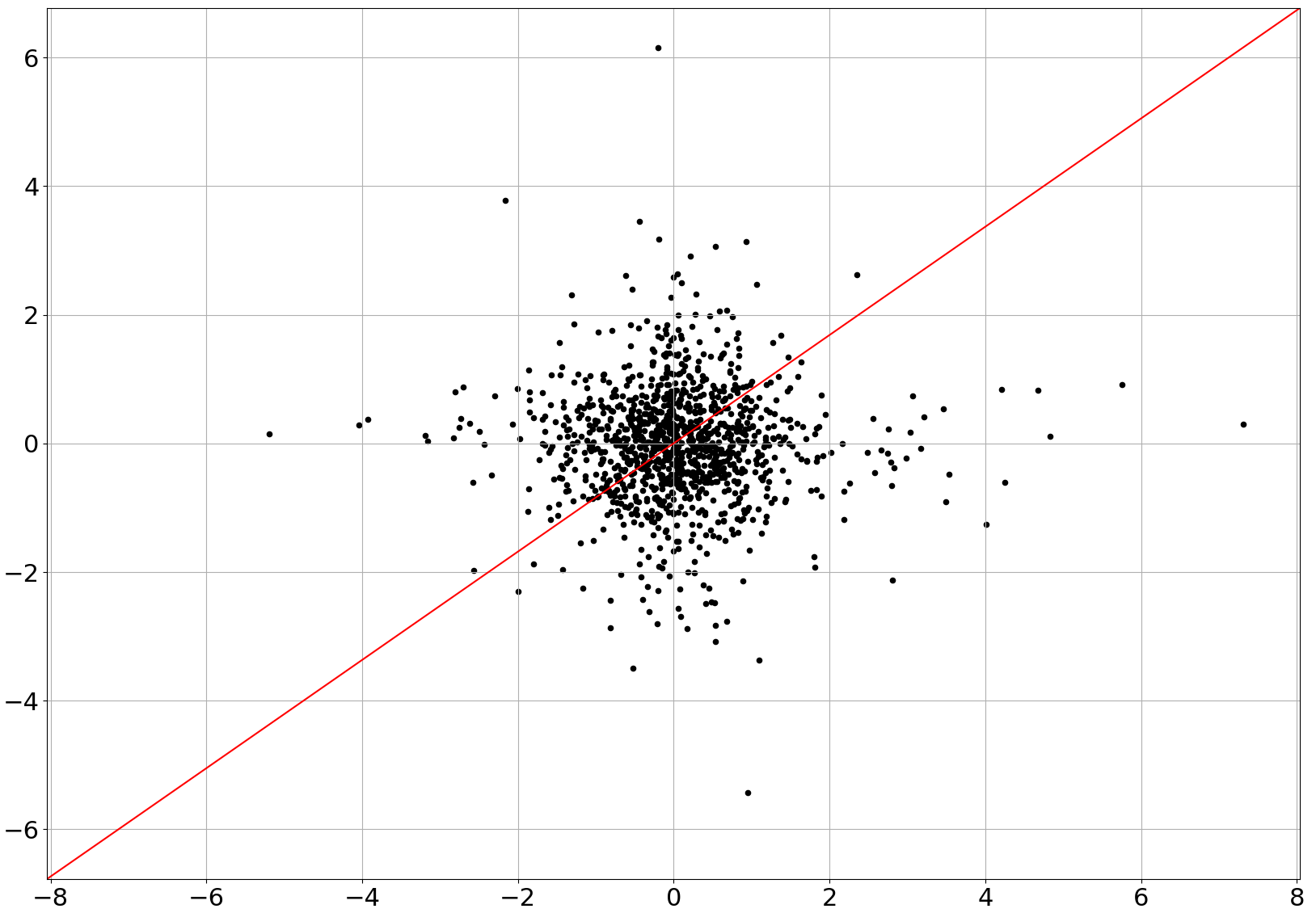}
 \caption{\small{\emph{PCA\_3}.}}
\end{subfigure}
\hfill
\caption{Scatter plots of the sovereign spread for Italy against Germany versus its top correlated covariates in the context of the \emph{DeepAR-Factors-GDELT-PCA} model, which are, respectively: (a) the first Nelson and Siegel term-structure factor (\emph{FACTOR\_1}), and the first three PCA factors extracted from GDELT data, i.e. (b) \emph{PCA\_1}, (c) \emph{PCA\_2}, and (d) \emph{PCA\_3}.}
\label{correlations_PCA-scatter}
\end{center}
\end{figure}

As a further alternative feature reduction method we have also run the Principal Component Analysis (PCA) over the GDELT variables \citep{Jollife2016}. PCA is a dimensionality-reduction method that is often used to reduce the dimensions of large data sets, by transforming a large set of variables into a smaller one that still contains the essential information characterizing the original data \citep{Jollife2016}. The results of a PCA are usually discussed in terms of component scores, sometimes called factor scores (the transformed variable values corresponding to a particular data point), and loadings (the weight by which each standardized original variable should be multiplied to get the component score) \citep{Jollife2016}. We have decided to use PCA with the intent to reduce the high number of correlated GDELT variables into a smaller set of ``important'' composite variables that are orthogonal to each other. Overall, PCA is used, when, like in our case, we want to retain the largest amount of variation in the original variables in the smallest number of variables possible. In principle, we could have estimated instead a factor analysis model \citep{factorsBartholomew20111}, a technique that, like PCA, allows reducing information in a larger number of variables. However, under the factor model approach the assumption is that there is a set of latent factors in a data set, and that each of the measured variables captures a part of one or more of those factors \citep{factorsVSPca-Schneeweiss1995105}. We believe that PCA is more suitable for our case, given that our aim is to build a set of artificial variables that capture the largest amount of variation in the original variables.


We have deployed PCA over the GDELT variables and selected the first three components only, i.e. the components that explain most of the variance in the news data set. These three components have been added as covariates to the three Nelson and Siegel term-structure factors, producing a model setting that we refer to as \emph{DeepAR-Factors-GDELT-PCA}. 
Figure \ref{correlations_PCA-scatter} reports the scatter plots of the target variable versus the four most highly correlated covariates in the context of this model. We notice that the first Nelson and Siegel term-structure factor, i.e. \emph{Factor 1}, is again, as expected, the top correlated feature, consistently also with what found in the feature selection step, see Figure \ref{correlations-scatter}. However \emph{Factor 1} is immediately followed by the first three PCA factors extracted from GDELT data, meaning that also the features coming from GDELT appear to be highly connected with the Italian sovereign spread.


For completeness, in our application we have also considered the estimation of the same models with GDELT variables alone, that is, without the three Nelson and Siegel term-structure factors. These are referred to as, respectively: \emph{DeepAR-GDELT}, namely the model that includes all the 51 preselected GDELT features; \emph{DeepAR-GDELT-hierarc}, including the GDELT features selected with hierarchical clustering (21 features); and \emph{DeepAR-GDELT-PCA}, considering the first three components of PCA deployed on all GDELT features.


\subsection{Model evaluation}

We compare the forecasting performance of the competing models by calculating a number of commonly used evaluation metrics and statistics \citep{Mehdiyev2016264}, such as the Root Mean Square Error ($RMSE$), the symmetric mean absolute percentage error (sMAPE), the $R^2$ and the \citet{Diebold1995} (DM) test statistics. 
To this end, we split the sample into two sub-samples of similar size: we take $T_0=586$ observations for estimation (from 2 March 2015 until 01 June 2017) and use the remaining 586 observations (from 02 June 2017 until 30 August 2019) for testing. 
Specifically, 
for $t=T_0+1,T_0+2,...,T-1$ the one-step-ahead forecast error of model $r$ and quantile $q$, using information up to time $t$ is calculated as:
\begin{equation}
 \hat{\epsilon}_{t+1}^{r} = \rho_q( \Delta Spread_{t + 1} - \widehat{\Delta Spread}_{t + 1,r} )
\end{equation}

where $\rho_q(z)$ is the so-called \textit{check function}, given by $\rho_q(z) = (q - I(z < 0))z$, and $\widehat{\Delta Spread}_{t + 1,r}$ is the forecast of the spread using the $r$-th model. 

\begin{figure}[!b]
\begin{center}
 \includegraphics[width=\textwidth]{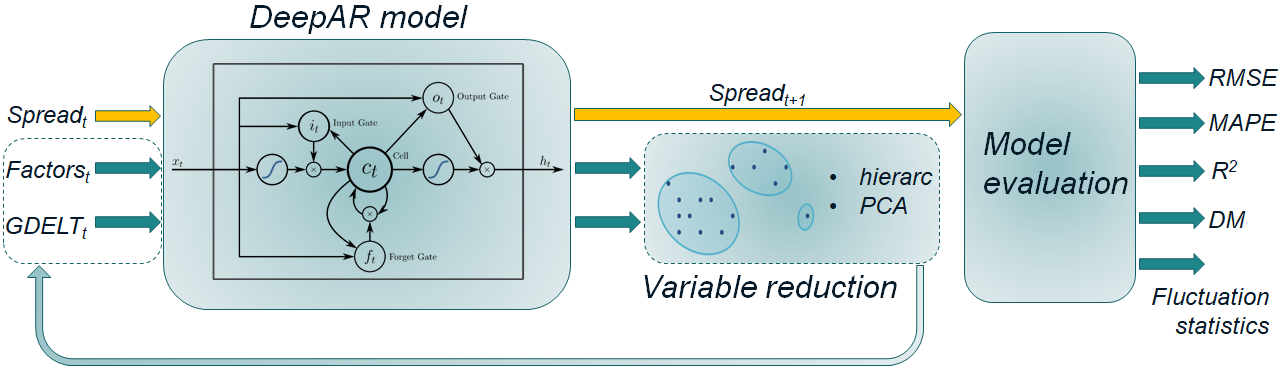}
	\caption{Graphical representation of the steps within the proposed algorithm pipeline.}
	\label{together-img}
\end{center}
\end{figure}

Our sample period covers a period of high political turmoil, as also evident from Figure \ref{target}. The large variations in the yield spread occurring particularly during the second half of the sample period hide important changes in the impact of our regressors on the spread over time, and undermine the validity of standard forecasting tests. As also pointed by \citet{Giacomini2010}, in the presence of structural instability, the relative performance of the two models may itself be time-varying, and thus averaging this evolution over time will result in a loss of information. By selecting the model that performs best on average over a particular historical sample, one may ignore the fact that the competing model produces more accurate forecasts when considering only the recent past. To account for such time variability, we employ a rolling window analysis, as already mentioned in Section \ref{deepar-meth}. Specifically, we re-estimate the model at each $t=T_0+1,T_0+2,...,T$ over a rolling window of $T_0$ days including data indexed $t -T_0,...,t-1$, 
where we set $T_0 =586$. For a statistical evaluation of the forecasting performance of our models, we adopt the \textit{Fluctuation test} proposed by \citet{Giacomini2010}. This test is suitable for comparing the out-of-sample forecasting performance of competing models in the presence of instabilities, as in our case. The Fluctuation test statistics consists of calculating the DM statistic over a rolling out-of-sample window of size $m$. When computing this test, we set the ratio between the rolling window length and the out-of-sample length, $\mu=m/(T-T_0)$, equal to $0.30$, as in \citep{Giacomini2010}, and use one lag in the calculation of the HAC estimator for the long-run variance. Finally, in selecting the relevant critical values for the Fluctuation test we consider a two-sided test and assume a nominal size of 5 percent (see Table I in \citep{Giacomini2010}).

Figure \ref{together-img} provides an overview of the proposed algorithm pipeline that includes the steps of the DeepAR model estimation, variable reduction, and model evaluation.

\section{Results} \label{experiments}

In this section we assess quantitatively the predicting power of the selected GDELT features over the Nelson-Siegel term structure factors using DeepAR.
%
%
Table \ref{DeepARloss} shows the minimised DeepAR loss function across different quantiles and different models calculated for the out-of-sample observations. It is interesting to observe the large reduction in the loss function when including the classical interest rates factor structure. When augmenting the model with both the factor structure and the GDELT variables we observe a further reduction in the loss function, particularly when adopting the \emph{DeepAR-Factors-GDELT-hierarc}	and \emph{DeepAR-Factors-GDELT-PCA} models. 
This suggests that news information extracted through data compression techniques over GDELT variables provides additional explanatory power over and above the classical Nelson and Siegel factors. For completeness, we also show the results of the models having only GDELT variables, i.e., respectively, \emph{DeepAR-GDELT}, \emph{DeepAR-GDELT-hierarc}, and \emph{DeepAR-GDELT-PCA}. These models outperform the pure autoregressive DeepAR, that is \emph{DeepAR-NoCov}, but they clearly underperform estimations that includes also the term-structure factors. In sum, the best performance is achieved when using jointly the classical factor structure and news information efficiently summarised in a few synthetic variables. Overall, the models obtain lower weighted quantile losses, thus the highest performance, for low quantiles, i.e 0.1.

\begin{table}[h!] \centering
\newcolumntype{Y}{>{\centering\arraybackslash}X}
\newcolumntype{V}{>{\centering\arraybackslash}p{5.1cm}}
\caption{DeepAR minimised loss function 
for 0.1, 0.3, 0.5, 0.7 and 0.9 quantiles.} \label{DeepARloss} \centering 
\begin{tabularx}{13.5cm}{l|YYYYY}
\hline\hline
\multicolumn{1}{r|}{Quantiles:}&\multicolumn{1}{c}{\bf{0.1}}&\multicolumn{1}{c}{\bf{0.3}}&\multicolumn{1}{c}{\bf{0.5}}&\multicolumn{1}{c}{\bf{0.7}}&\multicolumn{1}{c}{\bf{0.9}} \\ \hline
\multicolumn{1}{c|}{}&\multicolumn{5}{c}{}\\
\emph{DeepAR-NoCov}			 		&	410.96	&	443.88	&	442.48	&	456.76	&	432.33	\\
\emph{DeepAR-Factors}				 &	290.44	&	305.18	&	312.87	&	315.25	&	305.72	\\
\emph{DeepAR-Factors-GDELT}				&	249.54	&	263.29	&	269.67	&	276.13	&	276.83	\\
\emph{DeepAR-Factors-GDELT-hierarc}		&	\bf{202.55}	&	\bf{214.60}	&	\bf{218.32}	&	\bf{220.80}	&	\bf{220.53}	\\
\emph{DeepAR-Factors-GDELT-PCA} 		&	\bf{200.11}	&	\bf{214.02}	&	\bf{218.09}	&	\bf{219.81}	&	\bf{219.22}	\\
\emph{DeepAR-GDELT}						&	398.21	&	407.43	&	412.81	&	429.65	&	424.12	\\
\emph{DeepAR-GDELT-hierarc}				&	394.28	&	399.75	&	401.99	&	408.63	&	408.81	\\
\emph{DeepAR-GDELT-PCA}					&	396.87	&	397.09	&	406.01	&	407.25	&	407.83	\\
\hline\hline
\end{tabularx}
\bigskip
\end{table}

 
\newcolumntype{Y}{>{\centering\arraybackslash}X}
\newcolumntype{Z}{>{\arraybackslash}p{1.5cm}}
\begin{table}[t!] \centering
\caption{Out-of-sample forecast evaluation for the different \emph{DeepAR} models in terms of RMSE, sMAPE, and $R^2$ metrics.} \label{GroupResdeepar} \centering 
\begin{tabularx}{10.5cm}{l|YYY}
\hline\hline
\multicolumn{1}{r|}{Metrics:}&\multicolumn{1}{c}{\bf{RMSE}}&\multicolumn{1}{c}{\bf{sMAPE}}&\multicolumn{1}{c}{$\bf{R^2}$} \\ \hline
\multicolumn{1}{l|}{}&\multicolumn{3}{c}{}\\
\emph{DeepAR-NoCov}			 		 &	1.060	&	1.488	&	0.001	\\
\emph{DeepAR-Factors}				 &	0.833	&	1.080	&	0.246	\\
\emph{DeepAR-Factors-GDELT}				&	0.746	&	0.978	&	0.340	\\
\emph{DeepAR-Factors-GDELT-hierarc}		&	\bf{0.641}	&	\bf{0.847}	&	\bf{0.470}	\\
\emph{DeepAR-Factors-GDELT-PCA} 		&	\bf{0.640} &	\bf{0.845}	&	\bf{0.470}	\\
\emph{DeepAR-GDELT}						&	1.035	&	1.470	&	0.010	\\
\emph{DeepAR-GDELT-hierarc}				&	1.020	&	1.460	&	0.012	\\
\emph{DeepAR-GDELT-PCA}					&	1.008	&	1.455	&	0.012	\\
\hline\hline
\end{tabularx}
\end{table}

\begin{figure}[th!]
\centering
\begin{subfigure}[b]{0.79\textwidth}
 \centering
 \includegraphics[width=\textwidth]{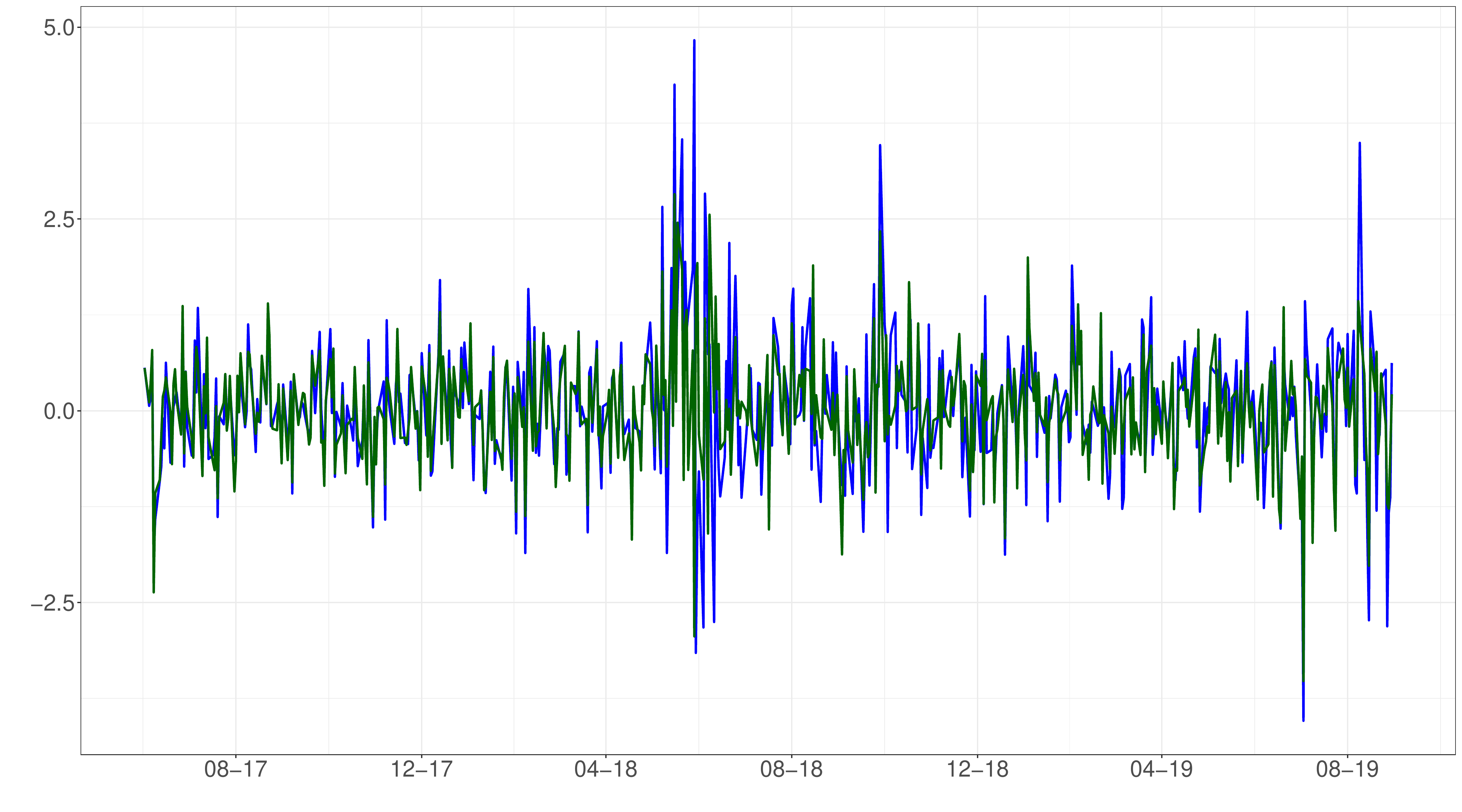}
\end{subfigure}
\vfill
\bigskip
\begin{subfigure}[b]{0.78\textwidth}
 \centering
 \includegraphics[width=\textwidth]{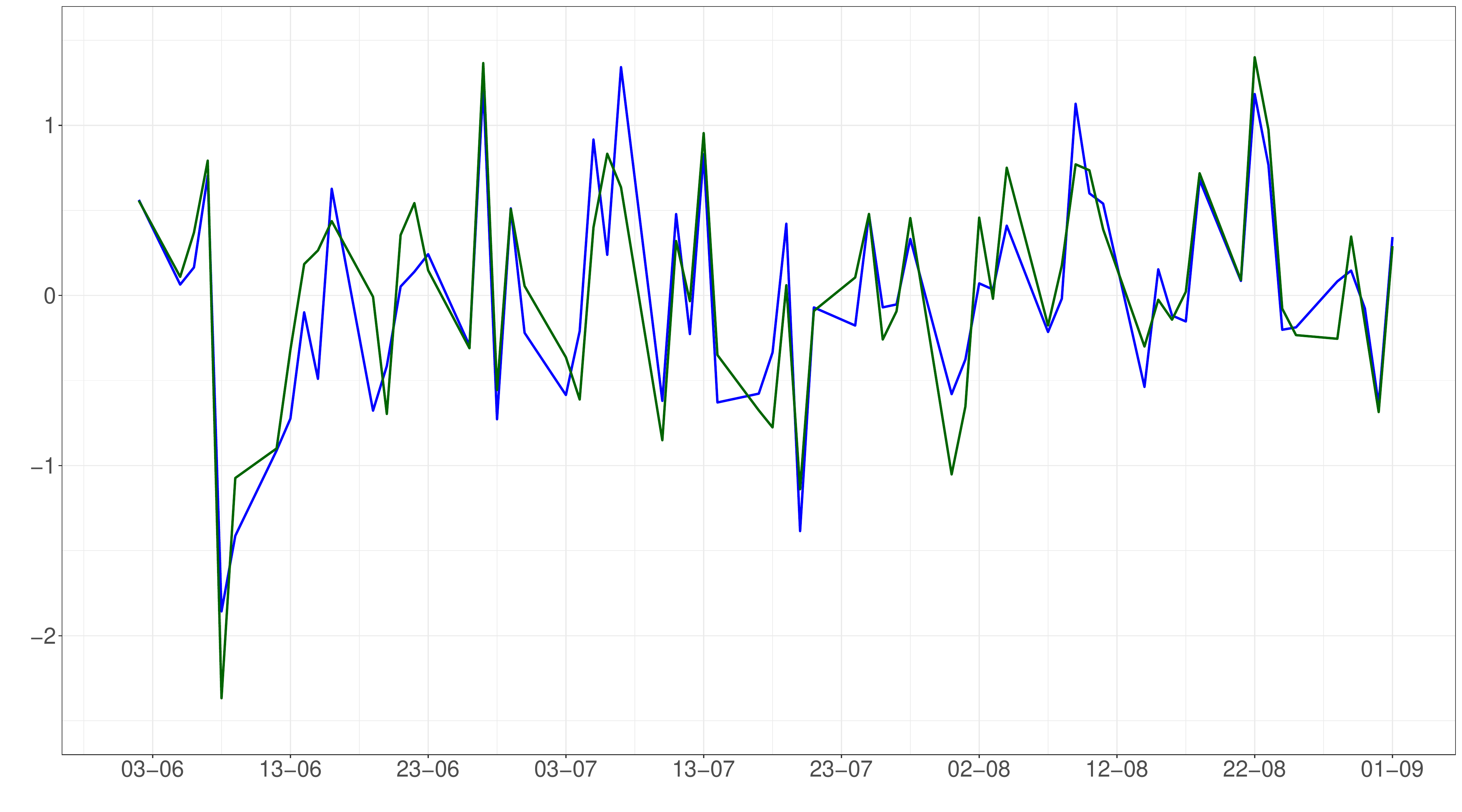} 
\end{subfigure}
\vfill
\caption{Median forecasts (green) for the \emph{DeepAR-Factors-GDELT-PCA} model and observations for the target series (blue) for the entire forecasting period (top) and for a zoom on the first 50 days (bottom).}
\label{Forecast-Prob_context-0}
\end{figure}

In Table \ref{GroupResdeepar} we also report the RMSE, sMAPE and $R^2$ focusing on the 0.5 quantile, i.e. on the median forecast, with the intent to zoom on the models' performance at the centre of the distribution. 
We note that, as expected, the three Nelson and Siegel term-structure factors are still very strong predictive variables. When these factors are added to the DeepAR specification, producing the \emph{DeepAR-Factors} model, the results get large improvements in terms of the RMSE, sMAPE and $R^2$ metrics. When we add the 51 GDELT variables to this model, we obtain a model setting that further improves consistently, pointing at the predicting power hidden in the selected GDELT variables. The best performances are obtained when we further reduce the considered GDELT variables, with the models \emph{DeepAR-Factors-GDELT-hierarc}	and \emph{DeepAR-Factors-GDELT-PCA}, resulting in a comparable, very high performance.

To get a qualitative outlook, in Figure \ref{Forecast-Prob_context-0} we report the observations for the target time series (blue line) together with the median forecast (dark green line) for the \emph{DeepAR-Factors-GDELT-PCA} model (top).
To better appreciate the differences between observed and predicted time series, at the bottom graph of Figure \ref{Forecast-Prob_context-0} we also report the same plot on a shorter time range, by zooming on the first 50 days of the testing period. 
A qualitative analysis of the figure suggests that the forecasting model does a very good job in capturing the dynamics of our target variable. 

\begin{figure}[ht!]
\begin{center}
 \includegraphics[width=0.49\linewidth]{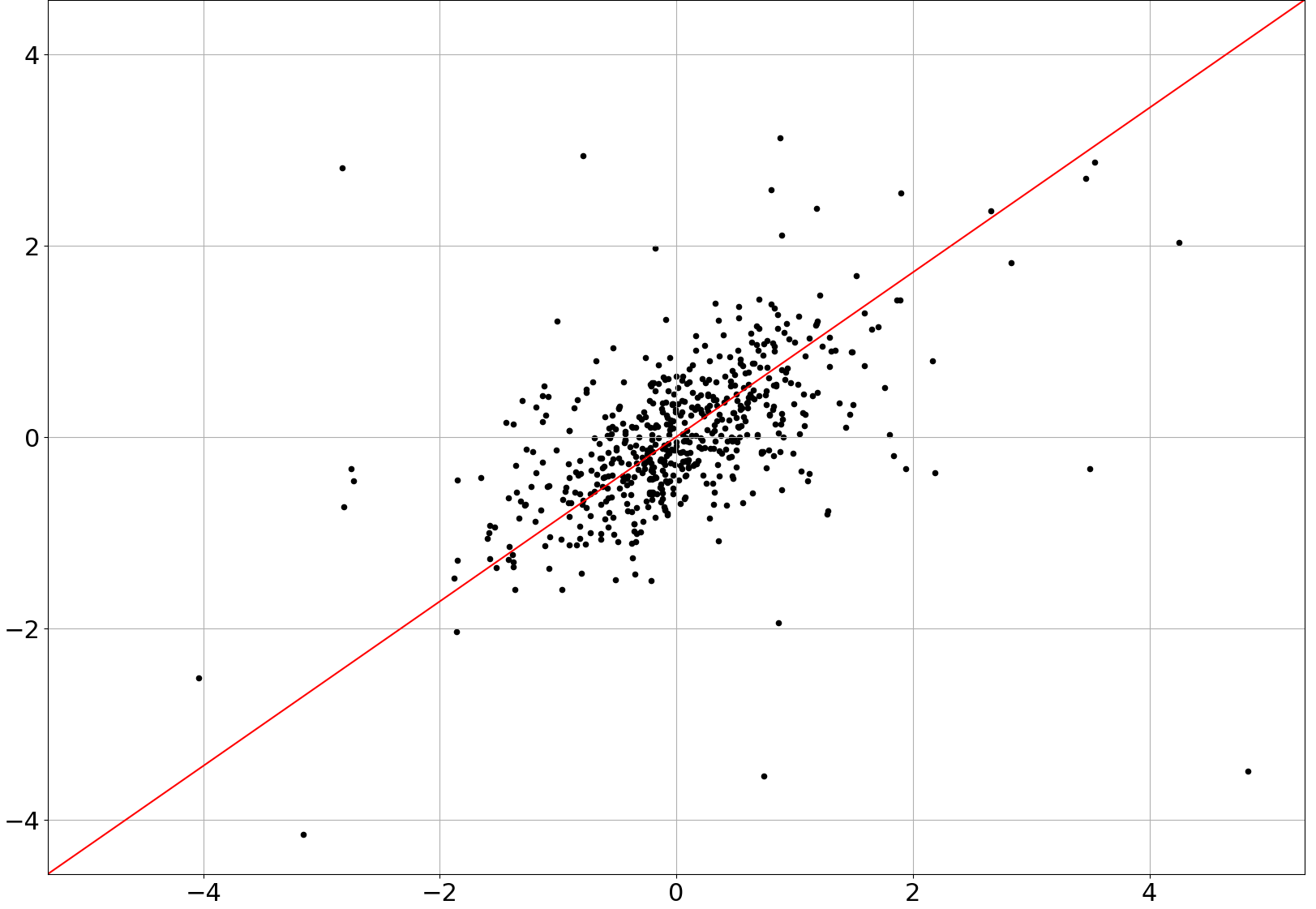}
 \includegraphics[width=0.49\linewidth]{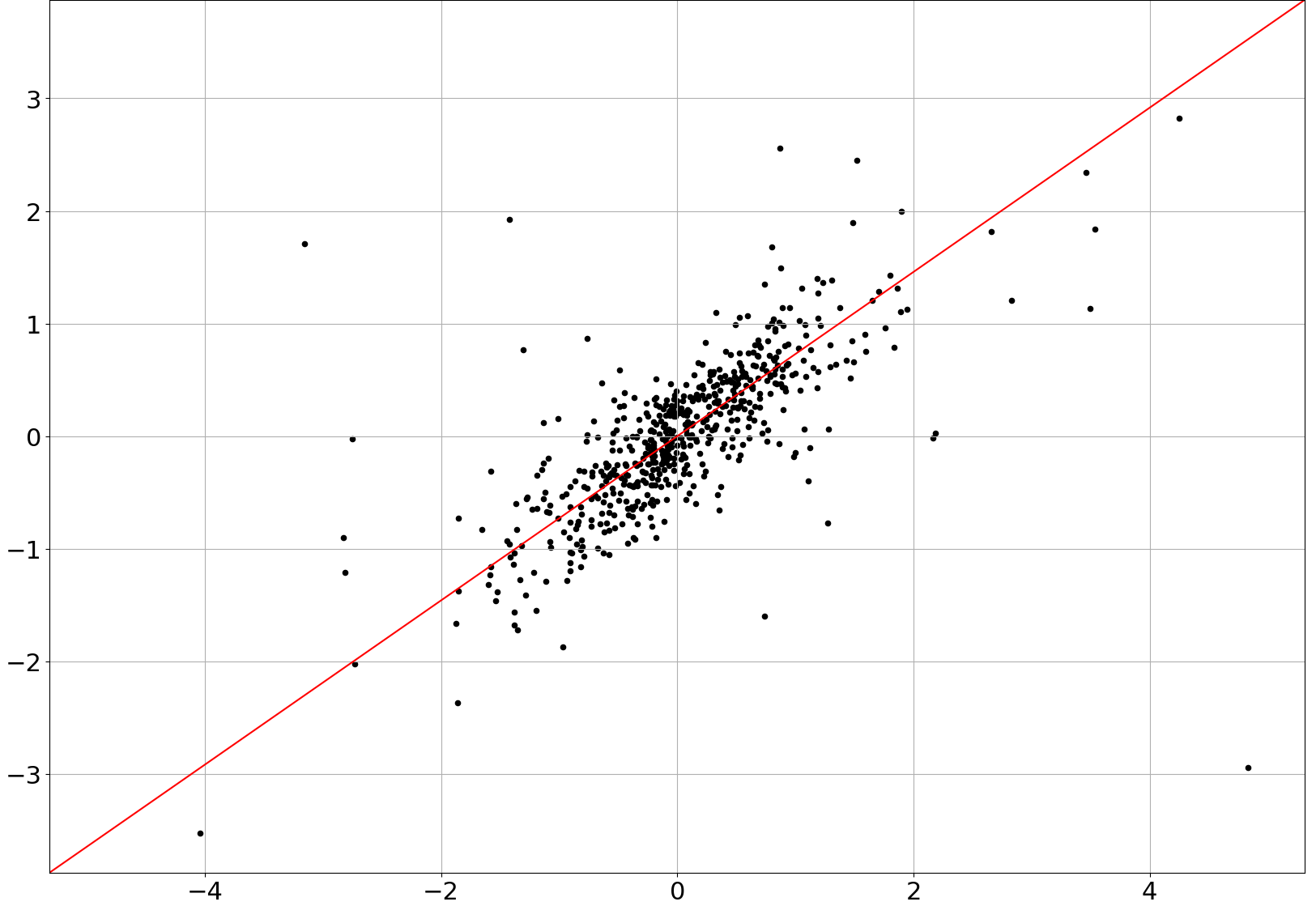}
	\caption{Scatter plot of the target variable versus the median out-of-sample forecasted points for \emph{DeepAR-Factors} (left) and for \emph{DeepAR-Factors-GDELT-PCA} (right).}
	\label{scatterplot}
\end{center}
\end{figure}

To better assess the improvements that we obtain when GDELT variables are considered jointly with the three Nelson and Siegel term-structure factors, in Figure \ref{scatterplot} we report the scatter plot of the target variable versus the median out-of-sample forecasted points for, respectively, the \emph{DeepAR-Factors} and \emph{DeepAR-Factors-GDELT-PCA} models. In both scatter plots we can observe that the points tend to concentrate along the main diagonal, underlining 
a good correlation between real and forecasted observations, and suggesting good quality of the forecasting results for both models. This is also confirmed by the high $R^2$ obtained for these models, equal to $0.246$ and $0.47$, respectively. However, we can clearly note that the points in the graph for the \emph{DeepAR-Factors-GDELT-PCA} are more densely concentrated along the diagonal, suggesting a higher degree of correlation between forecasted and real data, in line with the better forecasting metrics obtained by \emph{DeepAR-Factors-GDELT-PCA}.
 
\begin{figure}[t!]
\begin{center}
 \includegraphics[width=0.80\linewidth]{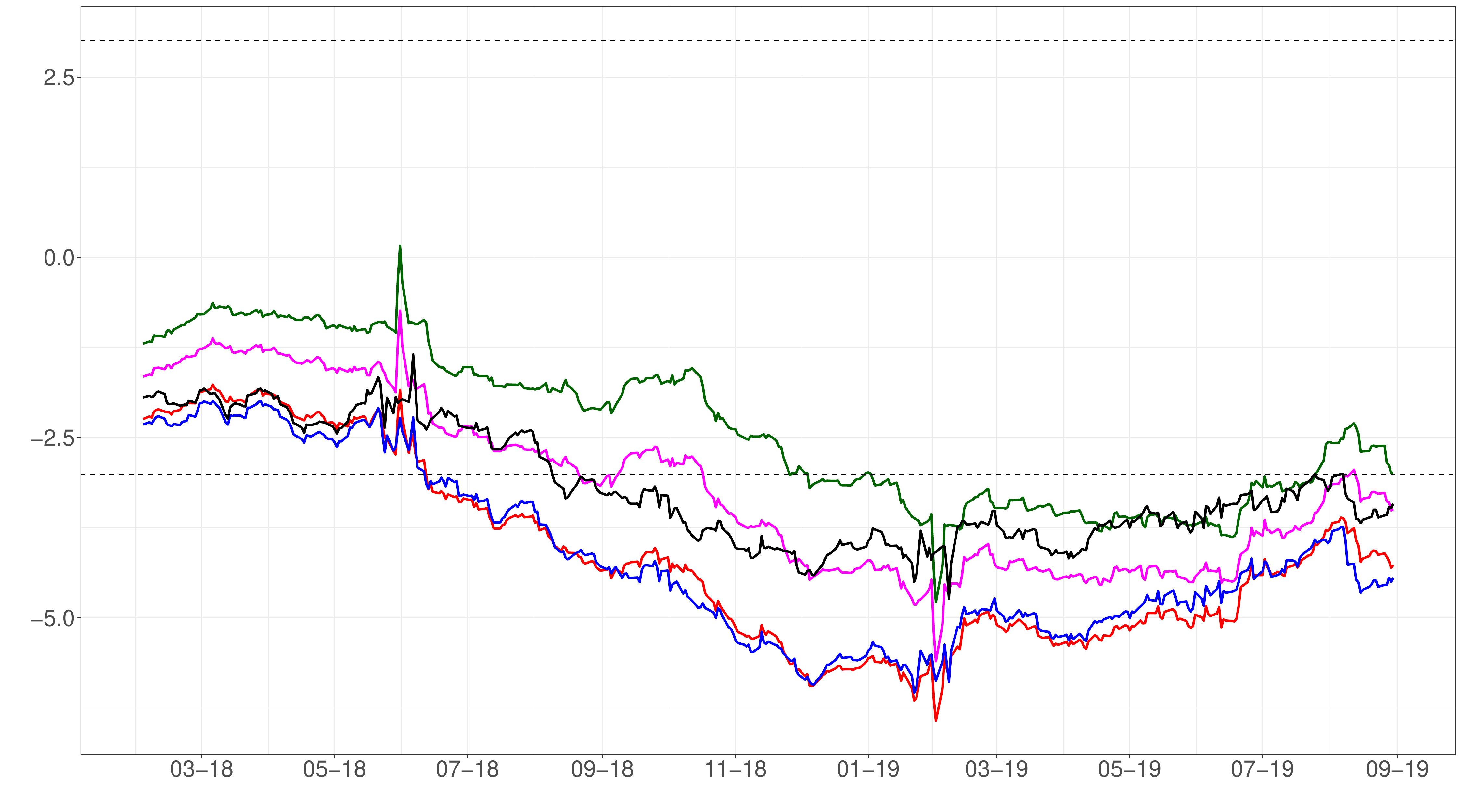}
 \caption{Fluctuation test statistic for \emph{DeepAR-Factors-GDELT-PCA} against \emph{DeepAR-Factors} for quantiles 0.1 (green), 0.3 (Magenta), 0.5 (red), 0.7 (blue) and 0.9 (black). The dashed line represents the critical value at the 5 percent significance level.}
	\label{fluct1}
\end{center}
\end{figure}

We now turn to a statistical comparison between the forecasting performance of these two models using the \citet{Giacomini2010} Fluctuation test statistic across different quantiles (Figure \ref{fluct1}). It is interesting to observe that for most of the testing period, the Fluctuation test is statistically significant at the 5 percent significance level, indicating the superiority of the news-augmented DeepAR model against the \emph{DeepAR-Factors} baseline. Interestingly, the improved performance varies across quantiles. While at lower quantiles (i.e., at 0.1) news information embedded in the GDELT factors starts to be important only at the end of 2018, the superiority of the news augmented model covers a relatively longer period of time when we forecast medium and medium-high quantiles (i.e., at 0.5 and 0.7). In these cases, the improvement in the performance of our \emph{DeepAR-Factors-GDELT-PCA} over the benchmark starts in June 2018, hence at the beginning of the Italian turbulent political period, and lasts till the end of our sample. Finally, the higher forecasting performance of our news augmented model for quantiles 0.3 and 0.9, starts only in August 2018.

\begin{figure}[t!]
\begin{center}
\includegraphics[width=0.80\linewidth]{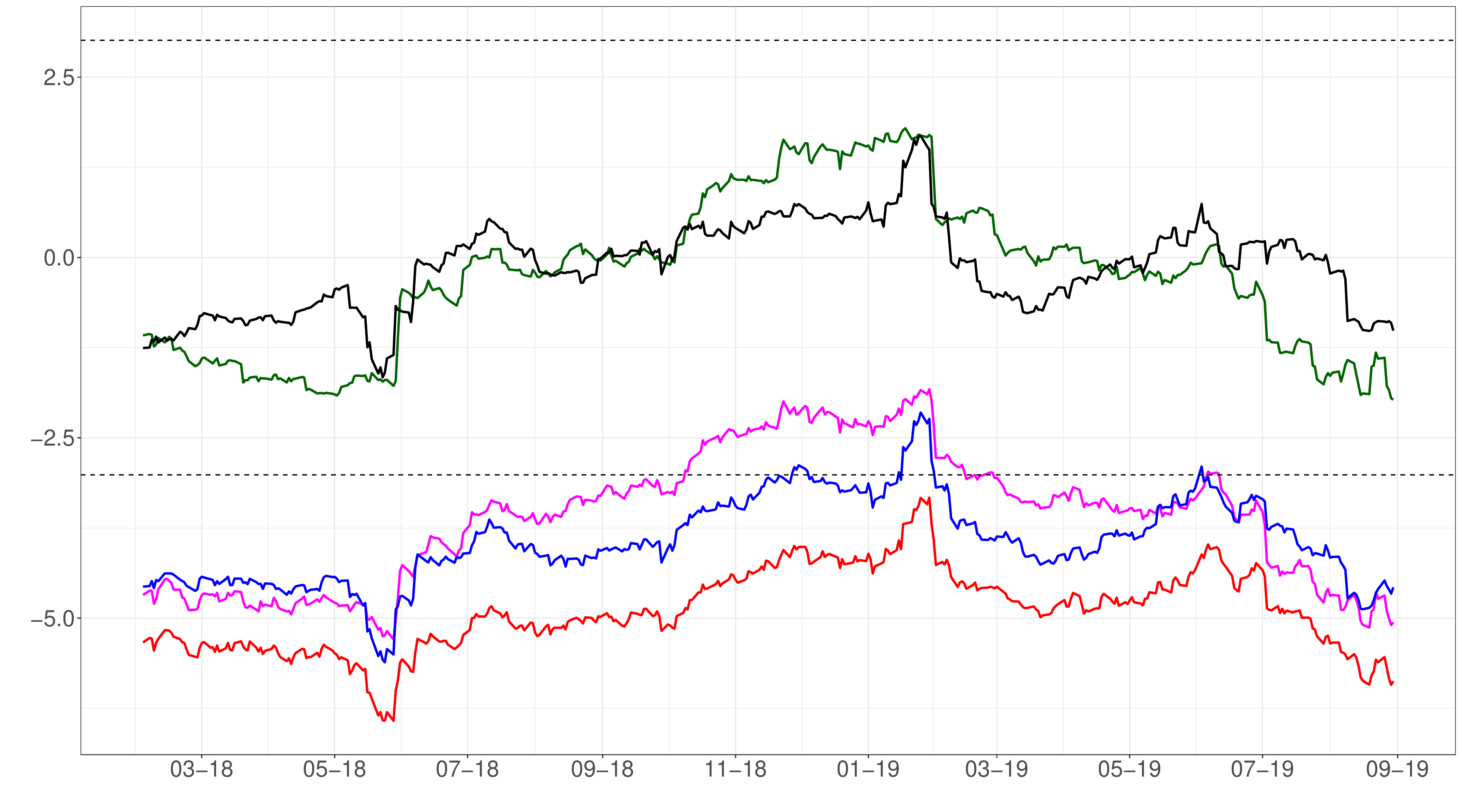}
\caption{Fluctuation test statistic for \emph{DeepAR-Factors-GDELT-PCA} against \emph{GB-Factors-GDELT-PCA} for quantiles 0.1 (green), 0.3 (Magenta), 0.5 (red), 0.7 (blue) and 0.9 (black).}	\label{fluct2}
\end{center}
\end{figure}

\begin{table}[!hb] \centering
\caption{Out-of-sample forecast evaluation for the different \emph{GB} models in terms of RMSE, sMAPE, and $R^2$ metrics.} \label{GroupResGB} \centering 
\begin{tabularx}{10.5cm}{l|YYY}
\hline\hline
\multicolumn{1}{c|}{Metrics:}&\multicolumn{1}{c}{\bf{RMSE}}&\multicolumn{1}{c}{\bf{sMAPE}}&\multicolumn{1}{c}{$\bf{R^2}$} \\ \hline
\multicolumn{1}{l|}{}&\multicolumn{3}{c}{}\\
\emph{GB-Factors}			&	0.878	&	1.863	&	0.004	\\
\emph{GB-Factors-GDELT}		&	0.876	&	1.813	&	0.000	\\
\emph{GB-Factors-GDELT-hierarc}&0.879	&	1.851	&	0.009	\\
\emph{GB-Factors-GDELT-PCA} &	0.876 &	1.842	& 0.000	\\
\emph{GB-GDELT}				&	0.875	&	1.814	&	0.001	\\
\emph{GB-GDELT-hierarc}		&	0.877	&	1.839	&	0.001	\\
\emph{GB-GDELT-PCA}			&	0.877	&	1.812	&	0.000	\\
\hline\hline
\end{tabularx}
\end{table}

As a final comparison, we have also checked whether the forecasting performance of the DeepAR model is superior to that of an alternative model such as the Gradient Boosting (GB) model \citep{Hastie2009,Natekin2013GBM}, often used for financial applications \citep{Bianchi_Buchner_Tamoni_2020,Gu_Kelly_Xiu_2020,Yang2020231,Liu2019851,Deng2019,Chang2018914}. GB is a well-known machine learning approach that produces a prediction by an ensemble of weak prediction models, typically decision trees, but that does not consider the time dependency in the target variable and covariates. It builds the model in a stage-wise fashion like other boosting methods do, and it generalizes them by allowing optimization of an arbitrary differentiable loss function \citep{Hastie2009,Natekin2013GBM}. 
%
The aim of this further analysis is to test whether the ability of DeepAR to account for serial dependence in the data enhances the out-of-sample performance relative to GB methods that are not designed to account for time dependence. 

In Figure \ref{fluct2} we use the Fluctuation test to compare the forecasting performance of the \emph{DeepAR-Factors-GDELT-PCA} model against the GB method considering the same set of explanatory variables (i.e., \emph{GB-Factors-GDELT-PCA}).
We refer to Appendix C for further details on the selection of the optimal GB parameters. 
Looking at this graph, we can note that accounting for serial dependence in \emph{DeepAR-Factors-GDELT-PCA} shows to be more relevant for quantiles belonging to the central part of the distribution of the dependent variable, rather than when focusing on extreme quantiles, which are more akin to random walk phenomena, given they represent rare events, hardly predictable, where accounting for the serial dependence in the data does not make a real difference. Focusing then on the 0.5 quantile, i.e. on the median forecast, we can see that the performance of \emph{DeepAR-Factors-GDELT-PCA} is indeed much superior with respect to that of \emph{GB-Factors-GDELT-PCA}, as shown in Table \ref{GroupResGB} which reports the RMSE, sMAPE and $R^2$ metrics for the different GB models considering the
various covariates combinations, that is: \emph{GB-Factors}, \emph{GB-Factors-GDELT}, \emph{GB-Factors-GDELT-hierarc}, \emph{GB-Factors-GDELT-PCA}, \emph{GB-GDELT}, \emph{GB-GDELT-hierarc}, and \emph{GB-GDELT-PCA}.

Looking at this table we can see that \emph{GB-Factors-GDELT-PCA} obtains an $RMSE = 0.876$, an $sMAPE = 1.842$, and an $R^2 = 0.000$, relative to a much better $RMSE = 0.64$, $sMAPE = 0.845$, and $R^2 = 0.47$ obtained by the corresponding \emph{DeepAR-Factors-GDELT-PCA} model (Table \ref{GroupResdeepar}), confirming the superiority of the DeepAR approach with respect to the 
analogous GB model with the same covariates setting.

From Table \ref{GroupResGB} we can also denote that 
the GB approach does not look being able at capturing the predictive patterns hidden within the news-based data. Indeed all the GB models including the GDELT features achieve comparable results which do not improve 
over the \emph{GB-Factors} model alone. This clearly suggests a weak predictive capability in this context of a machine learning approach such as the GB, which does not consider the temporal dependency in the target variable and covariates and further justifies the implementation of a more complex and yet comprehensive model as the employed DeepAR methodology.



\begin{figure}[!b]
\begin{center}
   \includegraphics[width=0.85\linewidth]{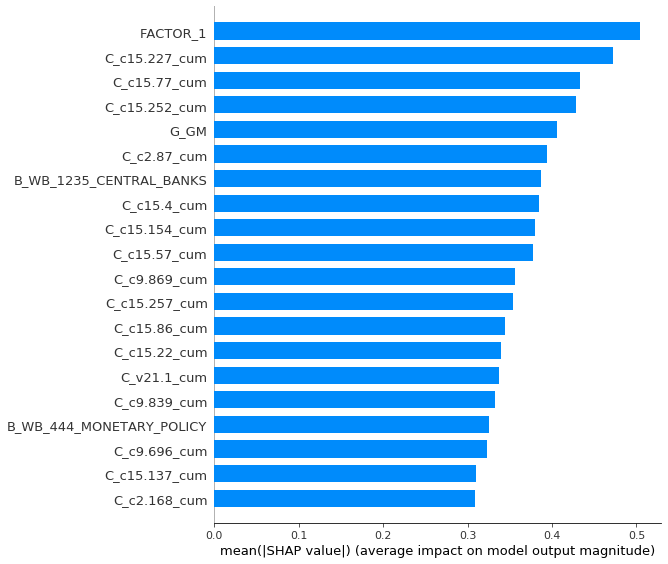}
	\caption{SHAP values bar plot: average impact of top features on model output magnitude.}
	\label{shap1}
\end{center}
\end{figure}

To improve the interpretability of the DeepAR model and explore the impact of the considered features on the model output,
we also perform the computation of the Shapley values \citep{NIPS2017_7062} of the model using the SHAP library available for Python \citep{lundberg2020local2global}. 
SHAP (SHapley Additive exPlanations) is a game-theory approach to explain the output of any machine learning model \citep{NIPS2017_7062}. It connects optimal credit allocation with local explanations using the classic Shapley values from game theory and their related extensions \citep{lundberg2020local2global}. 
In our case, we use the model agnostic KernelExplainer method, which is used to explain any function by using a specially-weighted local linear regression to estimate SHAP values for the considered model. 
For this analysis, we only focus on the model having the Nelson and Siegel term-structure factors and all GDELT variables as covariates, that is \emph{DeepAR-Factors-GDELT}. 
To get an overview of which features result to be the most important for the DeepAR model, in Figure \ref{shap1} we illustrate a standard bar plot with the mean absolute SHAP values of the top features over the data sample, sorting by importance from the most impactful features with respect to the model output to the worst ones. As expected, \emph{FACTOR\_1} has the largest impact, although this variable is immediately followed by other GDELT features generally referring to emotions, such as \textit{sadness}, \textit{diffidence}, \textit{stupefaction}, \textit{confidence}, \textit{distress}, \textit{anxiety} and \textit{happiness} (see Table \ref{GroupResQuantiles} in Appendix B for the complete mapping).
These results confirm the findings in previous literature that have shown media emotions to be relevant predictors for movements in spreads \citep{Apergis2016}, bond markets interest rates \citep{BEETSMA2013, Liu2014}, and stock prices \citep{tetlock2007}, and pointing at the explanatory and predictive power of these alternative variables in conjunction with classical determinants used for government credit spreads forecasts. 

\section{Conclusions} \label{conclusions} 

In this work we have exploited GDELT, the worldwide largest news-based database, to construct news-based financial indicators capturing topical and emotional content regarding the bond markets dynamics.
We have employed these variables to forecast the temporal evolution of the 10-year Italian interest rates spread against its German counterpart, during a period of market turbulence. We have adopted a DeepAR model tailored to 
account for possible non-linearity and time dependency in the data and to consider, as explanatory variables, the features extracted from GDELT along with classical Nelson and Siegel term structure factors. %
A dynamic statistical comparisons of the various considered models by means of the \citet{Giacomini2010} Fluctuations test, has indicated the significant contribution given by the GDELT variables across all quantiles of the predictive distribution. 
Furthermore, the results have shown the DeepAR methodology to be able to achieve high forecasting performance, capturing the predictive power of the news-based variables. A comparison with 
Gradient Boosting, another popular machine learning approach that does not consider the time dependency in the data, has confirmed the superiority of the DeepAR methodology that accounts instead for such time dependency in both target variable and covariates. DeepAR looks particularly suited to work in such a challenging context by obtaining very good out-of-sample performance scores.

This work represents one of the first to study the behaviour of government yield spreads in the presence of classical yield curve factors and information extracted from news, within a machine learning framework. 
This research could be extended in several directions. One possibility would be to widen the time span of the analysis to include the period of the economic crisis induced by the spread of the Covid-19 virus, which is still on-going at the time of this writing. 
We have decided on purpose to not consider the pandemics data into our analysis to avoid over-fitting of the model to this period, characterised by exceptional volatility, strong fluctuations in interest rates dynamics, and hard predictable behaviour. We believe that a satisfactory analysis of the long pandemics lapse requires an ad-hoc study with enough data to be undertaken a posteriori, in order to guarantee the due model reliability and out-of-sample accuracy. This will be the object of a separated study which we aim to develop in future work. 
An investigation of the model robustness to various adversarial attacks \citep{Adversial} maybe also a possible further extension of the present work. 

\section*{Acknowledgements}

The authors would like to thank the colleagues of the Centre for Advanced Studies at the Joint Research Centre of the European Commission for helpful guidance and support during the development of this research work.  We are grateful to the participants of the ``International Symposium on Forecasting'' for the numerous comments that significantly improved the paper.

%
%
\bibliography{main}


\section*{Appendix}

\subsection*{A. List of selected newspapers}

Below we report the list of outlets by country of origin (Italy) that have been selected for the analysis:
\newline
\textit{Il Sole 24 Ore, Borsa Italiana, Italia Oggi, Milano Finanza, Ansa, il Giornale, Finanza, Wall Street Italia, la Repubblica, Investire Oggi, Libero Quotidiano, il Messaggero Economia, il Fatto Quotidiano, il Corriere della Sera, Huffington Post Italy, La Stampa, trend-online.com, teleborsa, tradelink, il Tempo, finanza on-line, il Sussidiario.}

\subsection*{B. List of selected GDELT features}
Table \ref{GroupResQuantiles} provides the list of the selected GDELT features. The column indicated with the label: \emph{Hierarc}, indicates with a tick that the corresponding GDELT feature has been retained by the hierarchical clustering procedure and considered within the \emph{DeepAR-Factors-GDELT-hierarc} model.

\captionsetup[table]{skip=3pt} 
\begin{table}[!ht] \centering
\begin{footnotesize}
\caption{List of selected GDELT features. A tick in the \emph{Hierarc} column indicates that the corresponding GDELT feature has been retained by the hierarchical clustering procedure and considered within the \emph{DeepAR-Factors-GDELT-hierarc} model.} \label{GroupResQuantiles} \centering 
\begin{tabular}{l|ll|c}
\hline\hline
\multicolumn{1}{c|}{\emph{Themes}}&\multicolumn{2}{c|}{\bf{Taxonomy}}&\multicolumn{1}{c|}{\bf{Hierarc}}\\ \hline
\emph{INFLATION}				& 	\multicolumn{2}{l|}{\bf{World Bank (442)}}	&	\checkmark \\
\emph{MONETARY-POLICY}			& 	\multicolumn{2}{l|}{\bf{World Bank (444)}} & \checkmark \\
\emph{DEBT}			 		& 	\multicolumn{2}{l|}{\bf{World Bank (450)}} & \checkmark \\
\emph{CENTRAL-BANKS}				& 	\multicolumn{2}{l|}{\bf{World Bank (1235)}} & 	\\
\emph{CRISISLEXREC}				& 	\multicolumn{2}{l|}{\bf{Crisislex}} & \checkmark \\
\emph{LEADER}				 & 	\multicolumn{2}{l|}{\bf{GDELT}} &\\
\emph{GENERAL-GOVERNMENT}	 & 	\multicolumn{2}{l|}{\bf{GDELT}} &\\
\emph{USPEC-POLICY}				& 	\multicolumn{2}{l|}{\bf{GDELT}} & \checkmark\\
\emph{TAX-ECON-PRICE}		 & 	\multicolumn{2}{l|}{\bf{GDELT}} &\\
\emph{ECON-TAXATION} 		 & 	\multicolumn{2}{l|}{\bf{GDELT}} &\\
\emph{ECON-STOCKMARKET}		 & 	\multicolumn{2}{l|}{\bf{GDELT}} & \checkmark \\
\hline\hline
\multicolumn{1}{c|}{\emph{Dictionary Code}}&\multicolumn{1}{c}{\bf{Dictionary}}&\multicolumn{1}{c}{\bf{Dimension}}&\multicolumn{1}{|c}{\bf{Hierarc}}\\ \hline
\emph{2.168}					&	General Inquirer IV-Harvard	&	Payment loss	& \checkmark\\ 
\emph{2.42}					&	General Inquirer IV-Harvard	&	Decr & \\
\emph{2.87}					&	General Inquirer IV-Harvard &	Incr	& \\
\emph{8.1}					&	Regressive Imagery	&	Affection & \\
\emph{8.5}					&	Regressive Imagery	&	Glory	& \\
\emph{8.6}					&	Regressive Imagery	&	Positive Affect	& 	\\
\emph{8.7}					&	Regressive Imagery	&	Sadness	& 	\\
\emph{9.297}					&	Roget's Thesaurus	&	Recession	& \\
\emph{9.389}					&	Roget's Thesaurus	&	Pleasure & \checkmark \\
\emph{9.496}					&	Roget's Thesaurus	& Certainty	& \checkmark \\
\emph{9.696}					&	Roget's Thesaurus	&	Warning		& \checkmark \\
\emph{9.839}					&	Roget's Thesaurus	&	Credit & \checkmark \\
\emph{9.857}					&	Roget's Thesaurus	&	Affections & \\
\emph{9.869}					&	Roget's Thesaurus	&	Discontent	& \\
\emph{9.936}					&	Roget's Thesaurus	&	Hate	& 	\\
\emph{15.137}					&	WordNet Affect	&	Happiness	& \\
\emph{15.154}					&	WordNet Affect	&	Identification	& \\
\emph{15.168}					&	WordNet Affect	&	Joy-Pride		& \\
\emph{15.175}					&	WordNet Affect	&	Lost-Sorrow	& \\
\emph{15.198}					&	WordNet Affect	&	Negative-Fear	& \\
\emph{15.201}					&	WordNet Affect	&	Neutral-Unconcern & \\
\emph{15.203}					&	WordNet Affect	&	Optimism	& \checkmark \\
\emph{15.207}					&	WordNet Affect	&	Pessimism	& \\
\emph{15.215}					&	WordNet Affect	&	Preference & \\
\emph{15.22}					&	WordNet Affect	&	Anxiety	 & \\
\emph{15.226}					&	WordNet Affect	&	Reverence	& \checkmark \\
\emph{15.227}					&	WordNet Affect	&	Sadness	 & \checkmark \\
\emph{15.251}					&	WordNet Affect	&	Stir	& \checkmark \\
\emph{15.252}					&	WordNet Affect	&	Stupefaction	& \\
\emph{15.257}					&	WordNet Affect	&	Thing	& \\
\emph{15.26}					&	WordNet Affect	&	Approval	& \checkmark \\
\emph{15.4}					&	WordNet Affect	&	Affection & \checkmark \\
\emph{15.57}					&	WordNet Affect	&	Confidence	& \\
\emph{15.62}					&	WordNet Affect	&	Coolness	& \checkmark \\
\emph{15.77}					&	WordNet Affect	&	Diffidence	& \checkmark \\
\emph{15.86}					&	WordNet Affect	&	Distress	& \checkmark \\
\emph{v19.2}					&	Affective Norms for English Words	&	Arousal	& 	\\
\emph{v19.3}					&	Affective Norms for English Words	&	Dominance	& 	\\
\emph{v21.1}					&	Hedonometer	&	Happiness score & 	\checkmark \\
\hline\hline
\multicolumn{1}{c|}{\emph{Location Code}}&\multicolumn{2}{c}{\bf{Country}} &\multicolumn{1}{|c|}{\bf{Hierarc}}\\ \hline
\emph{GM}				&	Germany		&	\\

\hline\hline
\end{tabular}
\end{footnotesize}
\end{table}

\subsection*{C. Selection of GB optimal parameters}

The main parameters to optimize in our GB model are the \textit{maximum tree depth}, which indicates the maximum possible depth of a tree in the model and is used to control over-fitting, as higher depth will allow model to learn relations very specific to a particular sample; and the \textit{learning rate}, which determines the impact of each tree on the final outcome of the GB model. GB works by starting with an initial estimate which is updated using the output of each tree; the learning rate parameter controls then the magnitude of this change in the estimates. Therefore, lower values are generally preferred as they make the model robust to the specific characteristics of the tree and thus allowing it to generalize well. However lower values would require higher number of trees to model all the relations and will be computationally expensive.
To determine the optimal parameter values of our GB model, we have used a 10-fold cross-validation together with a grid search (or parameter sweep) procedure~\citep{gridsearch}. Grid search involves an exhaustive searching through a manually specified subset of the hyperparameter space of the learning algorithm, guided by some performance metric (like in our case minimizing the mean squared error).

To explore the hyperparameters space looking for optimal values of these parameters, the grid search procedure has tested the GB model with values going from 1 to 10 (by steps of 2) for the maximum tree depth parameter and from 0.01 to 0.99 (by steps of 0.20, including 0.99) for the learning rate parameter. The best parameter values with respect to the mean squared error are produced as output.
%
Although grid search does not provide an absolute guarantee that it will find the global optimal parameter values, in practice we have found it to work quite well, despite to be quite computationally expensive. In general grid searching is a widely adopted and accepted procedure for this type of tuning tasks \citep{gridsearch}.

\end{document}